\documentclass[10pt,twocolumn,letterpaper,pagebackref,breaklinks,colorlinks,allcolors=iccvblue]{article}

\usepackage[pagenumbers]{iccv} 
\usepackage{hyperref}
\usepackage{adjustbox}
\usepackage{graphicx}
\usepackage{caption}
\usepackage{subcaption}
\usepackage{printlen}
\usepackage{float}
\usepackage{wrapfig}
\usepackage{multirow}
\usepackage{booktabs}
\usepackage{pifont}
\usepackage{makecell}
\usepackage{nicefrac}
\usepackage{url}
\usepackage{stmaryrd}
\usepackage{verbatim}
\usepackage{mathabx}
\usepackage{algorithm}
\usepackage{algpseudocode}
\usepackage{amsfonts}
\usepackage{microtype}
\usepackage{pgf}
\usepackage{pgfplots}
\pgfplotsset{compat=1.18} 
\usepackage{tabularx}
\usepackage{ifthen}
\usepackage{soul}
\usepackage{tikz}
\usetikzlibrary{angles,quotes,3d,math,arrows.meta,calc,positioning,fit,backgrounds,decorations.pathreplacing,calligraphy,shapes,shapes.multipart}
\usepackage{diagbox}
\usepackage{amsmath}

\newcommand{\fflow}[1]{\mathbf{f}_{\ifthenelse{\equal{#1}{}}{\rightarrow}{#1}{}}}

\makeatletter
\newcommand{\xleftrightarrow}[2][]{\ext@arrow 3359\leftrightarrowfill@{#1}{#2}}
\newcommand{\xdashrightarrow}[2][]{\ext@arrow 0359\rightarrowfill@@{#1}{#2}}
\newcommand{\xdashleftarrow}[2][]{\ext@arrow 3095\leftarrowfill@@{#1}{#2}}
\newcommand{\xdashleftrightarrow}[2][]{\ext@arrow 3359\leftrightarrowfill@@{#1}{#2}}
\def\rightarrowfill@@{\arrowfill@@\relax\relbar\rightarrow}
\def\leftarrowfill@@{\arrowfill@@\leftarrow\relbar\relax}
\def\leftrightarrowfill@@{\arrowfill@@\leftarrow\relbar\rightarrow}
\def\arrowfill@@#1#2#3#4{%
  $\m@th\thickmuskip0mu\medmuskip\thickmuskip\thinmuskip\thickmuskip
   \relax#4#1
   \xleaders\hbox{$#4#2$}\hfill
   #3$%
}
\makeatother

\newcommand{\remove}[1]{{\color{ourorange}#1}}

\renewcommand{\remove}[1]{}

\definecolor{ourgreen}{RGB}{46, 204, 113}
\definecolor{ourgreenborder}{RGB}{39, 174, 96}
\definecolor{ourblue}{RGB}{52, 152, 219}
\definecolor{ourblueborder}{RGB}{41, 128, 185}
\definecolor{ourorange}{RGB}{230, 126, 34}
\definecolor{ourorangeborder}{RGB}{211, 84, 0}
\definecolor{ourred}{RGB}{231, 76, 60}
\definecolor{ourredborder}{RGB}{192, 57, 43}
\definecolor{ouryellow}{RGB}{241, 196, 15}
\definecolor{ouryellowborder}{RGB}{243, 156, 18}
\definecolor{ourpurple}{RGB}{155, 89, 182}
\definecolor{ourpurpleborder}{RGB}{142, 68, 173}
\definecolor{ourturquoise}{RGB}{26, 188, 156}
\definecolor{ourturquoiseborder}{RGB}{22, 160, 133}
\definecolor{ourturquoise}{RGB}{26, 188, 156}
\definecolor{ourturquoiseborder}{RGB}{22, 160, 133}
\definecolor{ourwhite}{RGB}{236, 240, 241}
\definecolor{ourwhiteborder}{RGB}{189, 195, 199}
\definecolor{ourgray}{RGB}{149, 165, 166}
\definecolor{ourgrayborder}{RGB}{127, 140, 141}

\definecolor{ourwhite2}{RGB}{246, 247, 248}

\definecolor{ourhighlightcolor}{RGB}{46, 204, 113}

\newcolumntype{H}{>{\setbox0=\hbox\bgroup}c<{\egroup}@{}}

\usepackage{graphicx}
\usepackage{amsmath}
\usepackage{amssymb}
\usepackage{booktabs}
\usepackage{times}
\usepackage{microtype}
\usepackage{epsfig}
\usepackage{caption}
\usepackage{float}
\usepackage{placeins}
\usepackage{color}
\usepackage{stfloats}
\usepackage{enumitem}
\usepackage{tabularx}
\usepackage{xstring}
\usepackage{multirow}
\usepackage{xspace}
\usepackage{url}
\usepackage{subcaption}
\usepackage{arydshln}
\usepackage{bbm}
\usepackage[hang,flushmargin]{footmisc}
\usepackage{pifont}
\usepackage{wrapfig}
\usepackage{lipsum}
\usepackage{comment}
\usepackage{listings}
\usepackage{amsmath}
\usepackage{amssymb}
\usepackage{makecell}
\usepackage{pgfplots}
\usepackage{tikz}
\usepackage{arydshln}
\usepackage{bibunits}
\usepackage{xfrac}

\usepackage[capitalize]{cleveref}
\crefname{section}{Sec.}{Secs.}
\Crefname{section}{Section}{Sections}
\Crefname{table}{Table}{Tables}
\crefname{table}{Tab.}{Tabs.}

\usepackage{wrapfig}

\newcommand{\fidbaseline}[1]{%
  \text{#1}\textsubscript{\phantom{\textcolor{green}{\textnormal{$\downarrow9.75$}}}}%
}
\newcommand{\fidworsened}[2]{%
  #1$_{\color{ourorange}\blacktriangleup #2}$%
}
\newcommand{\fidimproved}[2]{%
  #1$_{\color{ourgreen}\blacktriangledown #2}$%
}

\usepackage{algorithm}
\usepackage{algpseudocode} 
\usepackage{amsmath}      
\usepackage{amsfonts}     
\usepackage{amssymb}      
\usepackage{comment}      

\usepackage[utf8]{inputenc}  
\usepackage{pgf}
\usepackage{amssymb}
\usepackage{pifont}
\usepackage{graphicx} 
\usepackage{xparse} 
\usepackage{scalerel} 
\usepackage[accsupp]{axessibility}
\NewDocumentCommand\emojione{O{1} O{}}{%
  \IfValueTF{#2}{%
    \includegraphics[width=#2]{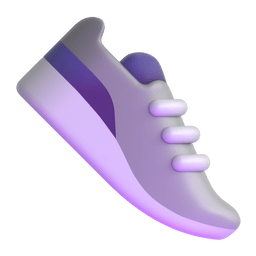}%
  }{%
    \scalerel*{\includegraphics[scale=#1]{fig/sneaker3.png}}{X}%
  }%
}
\newcommand{\methodname}{TREAD}

\newcommand{\route}{\textbf{r}}

\newcommand{\network}{D_\theta}

\newcommand{\methodnameoutsidetab}[4]{$\text{#1-#2}_{+ \textbf{\methodname#3} \text{ #4}}$}

\newcommand\rurl[1]{%
  \href{https://#1}{\nolinkurl{#1}}%
}
\definecolor{iccvblue}{rgb}{0.21,0.49,0.74}
\PassOptionsToPackage{pagebackref,breaklinks,colorlinks,allcolors=iccvblue}{hyperref}

\title{TREAD: Token Routing for Efficient Architecture-agnostic Diffusion Training}

\author{
    Felix Krause\hspace{0.8em}Timy Phan\hspace{0.8em}Ming Gui\hspace{0.8em}
    Stefan Andreas Baumann\hspace{0.8em}Vincent Tao Hu\hspace{0.8em}Björn Ommer\\[1ex]
    CompVis @ LMU Munich, \quad Munich Center for Machine Learning (MCML)\\
}

\begin{document}
\maketitle
\begin{abstract}
Diffusion models have emerged as the mainstream approach for visual generation. However, these models typically suffer from sample inefficiency and high training costs. Consequently, methods for efficient finetuning, inference and personalization were quickly adopted by the community. However, training these models in the first place remains very costly. While several recent approaches including masking, distillation, and architectural modifications have been proposed to improve training efficiency, each of these methods comes with a tradeoff: they achieve enhanced performance at the expense of increased computational cost or vice versa. In contrast, this work aims to improve training efficiency as well as generative performance at the same time through routes that act as a transport mechanism for randomly selected tokens from early layers to deeper layers of the model. Our method is not limited to the common transformer-based model - it can also be applied to state-space models and achieves this without architectural modifications or additional parameters. Finally, we show that \methodname{} reduces computational cost and simultaneously boosts model performance on the standard ImageNet-256 benchmark in class-conditional synthesis. Both of these benefits multiply to a convergence speedup of \textbf{14$\times$} at 400K training iterations compared to DiT and \textbf{37$\times$} compared to the best benchmark performance of DiT at 7M training iterations. Furthermore, we achieve a competitive FID of \textbf{2.09} in a guided and \textbf{3.93} in an unguided setting, which improves upon the DiT, without architectural changes. \newline 
Project Page: \mbox{\href{https://compvis.github.io/tread}{https://compvis.github.io/tread}}
\end{abstract}

\vspace{2mm}
\section{Introduction}

\begin{figure}[ht]
    \centering
    \adjustbox{max width=\linewidth}{
    \includegraphics[width=\linewidth]{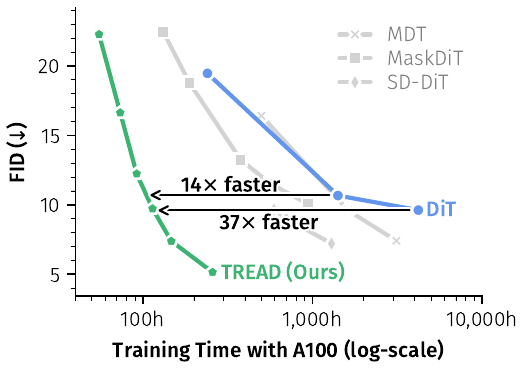}
    }
    \caption{We introduce TREAD, a training strategy that enables substantially more efficient training of token-based diffusion backbones. Applied to the standard backbone DiT~\cite{dit_peebles2022scalable}, we achieve a 14/37$\times$ training speed increase w.r.t. unguided FID while also converging to better generation quality.}
    \label{fig:fid_steps_scatter}
    \vspace{-3mm}
\end{figure}

\begin{figure*}[ht]
    \centering
    \adjustbox{max width=\textwidth}{
    \includegraphics[width=\textwidth]{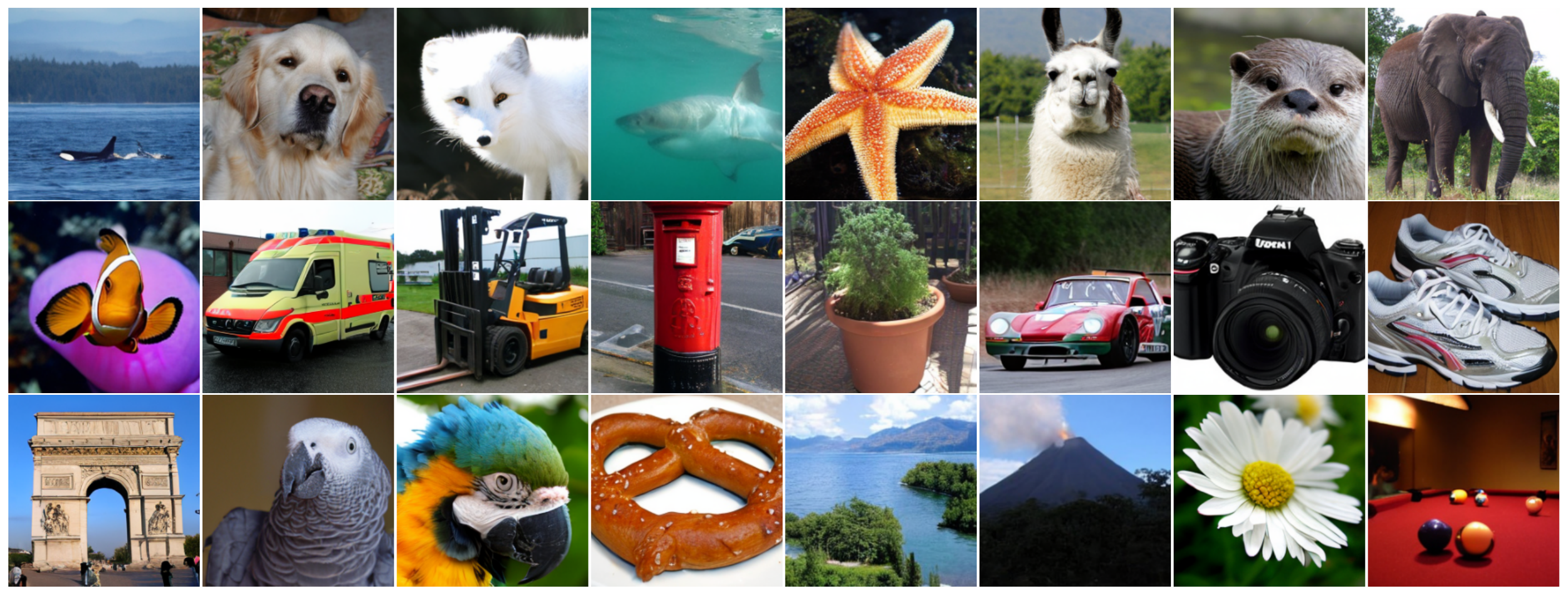}
    }
    \caption{\textbf{Selected samples from ImageNet-256} generated with a \methodnameoutsidetab{DiT}{XL/2}{}{} using a guidance weight of $\omega=3.5$.}
    \label{fig:samples}
\end{figure*}

In recent years, diffusion models \cite{sohl2015deep, ho2020denoising, rombach2022high_latentdiffusion_ldm} have become a powerful generative technique for image synthesis \cite{rombach2022high_latentdiffusion_ldm}. They have also been successfully extended to the 3D~\cite{poole2023dreamfusion} and video domains~\cite{blattmann2023align_videoldm}. While diffusion models avoid some training challenges faced by their predecessors, such as GANs \cite{goodfellow2020generative}, which are inherently unstable during training, they incur high costs \cite{liang2024scalinglawsdiffusiontransformers} due to slow convergence and sample inefficiency \cite{mei2023deepnetworksdenoisingalgorithms}. Currently, the Diffusion Transformer (DiT) \cite{dit_peebles2022scalable} is the main backbone when training diffusion models, building on the established Transformer architecture \cite{vaswani2017attention}. However, the Transformer architecture has computational cost limitations, as it scales quadratically with token length and converges slowly, further increasing the cost of training diffusion models. Training a DiT on standard benchmarks alone requires thousands of A100 GPU hours without reaching convergence, and text-to-image models demand even more resources. Stable Diffusion \cite{rombach2022high_latentdiffusion_ldm}, for instance, required about 150{,}000 A100 GPU hours. 
Despite ongoing efforts to reduce computational demands through improved infrastructure, implementation, and hardware, democratizing the training of diffusion models remains a distant goal. Methods such as LoRA \cite{hu2022lora} for parameter-efficient fine-tuning, caching strategies \cite{wimbauer2024cache, ma2024deepcache} to accelerate inference, and approaches for personalization \cite{gal2022image, hulfm,ruiz2023dreambooth, baumann2024continuous} have made these aspects more accessible. However, full-scale training remains extremely costly, limiting broad participation in the development and improvement of diffusion models.
Several works have proposed methods to accelerate training convergence utilizing external self-supervised features~\cite{hu2023self,yu2024repa,fuest2024diffusion}, improved flow-based theory~\cite{lipman2022flow,gui2024depthfm,schusterbauer2024boosting}, or optimized data combinations~\cite{tong2023improving, mixup}. 
Another approach to reducing computational resource requirements is \emph{masking}, where a subset of the available information is used during training. This subset can be selected either randomly \cite{zheng2023fast_maskdit, Gao_2023_ICCV} or based on learned heuristics \cite{wang2024attentiondriventrainingfreeefficiencyenhancement, meng2022adavit}, while the remaining information is permanently lost \cite{devlin2018bert, he2022masked_mae}. 
Beyond masking, computational efficiency can also be improved by controlling the flow of information during training. Instead of discarding information, it can be selectively directed to specific computational blocks, allowing tokens to skip computations and reducing the overall workload. This process is known as \emph{routing} \cite{raposo2024mixture}.

These methods offer two potential advantages: \textbf{i)} reducing compute requirements by processing fewer tokens \cite{zheng2023fast_maskdit, gupta2022maskvitmaskedvisualpretraining}, and \textbf{ii)} enhancing convergence by encouraging contextual relation learning \cite{assran2023selfsupervisedlearningimagesjointembedding, Gao_2023_ICCV}, thus addressing the lacking data efficiency and lowering computational demands. While masking has been extensively studied for both benefits and is a well-established technique in self-supervised~\cite{assran2023selfsupervisedlearningimagesjointembedding} and representation learning \cite{he2022masked_mae}, routing has primarily been considered from the viewpoint of compute reduction \cite{raposo2024mixture, he2024matters}.

This work explores the routing technique for diffusion models using a route that passes tokens from one layer to another layer deeper in the network. The loss of information is only temporary, as information will be reintroduced after the route concludes. We do this by defining a route with a start and end layer and a fixed ratio with which tokens are randomly selected for routing. This is done only during training and without any dynamic adaptations based on iterations or timesteps. Furthermore, we test a variety of configurations and derive clear guidelines for applying our routing mechanism to achieve substantial improvements in both qualitative performance and efficiency.
Our main contributions can be summarized as follows:

\begin{itemize}
    \item A novel training strategy for token-based sequential models that utilizes a token transport mechanism termed \emph{routing}, which directs tokens from early into later network layer.
    \item An evaluation of the underlying routing mechanism, its effectiveness, and the derivation of empirically validated guidelines for its application.
    \item Significant improvements in both convergence per step and cost per step on the ImageNet-256 benchmark, leading to a substantial reduction in computational cost. This enables the training of state-of-the-art competitor models at a fraction of the cost without requiring architectural modifications.
    \item Finally, we reach an FID of \textbf{2.09} which improves upon the standard DiT training and results in a \textbf{37$\times$} speedup.
\end{itemize}

\section{Related Work}

\paragraph{Diffusion Models and Efficient Generative Backbones.}
Score-based generative models \cite{song2019generative, song2020improved}, like notably DDPMs \cite{ho2020denoising}, now lead image synthesis. They define a forward SDE that progressively adds Gaussian noise, mapping data to a standard Gaussian. An iterative denoising process is used to reverse this transformation and generate samples. A key improvement in the efficiency of early diffusion models was to train in a compressed latent space \cite{rombach2022high_latentdiffusion_ldm}.
Building on the foundation of score‑based models, early diffusion methods \cite{ho2020denoising, dhariwal2021diffusion, rombach2022high_latentdiffusion_ldm} used the UNet backbone \cite{unet}, while more recent token‑based architectures like DiT \cite{dit_peebles2022scalable} and its variants \cite{zheng2023fast_maskdit, esser2024scalingrectifiedflowtransformers, chen2023pixartalphafasttrainingdiffusion} have gained preference, despite their quadratic complexity in the number of tokens.
To address computational challenges associated with token-based models, recent work introduces caching for UNet‑based diffusion \cite{ma2023deepcache} or DiT‑based models \cite{ma2024learningtocacheacceleratingdiffusiontransformer}. Unlike these approaches, our method instead applies a training‑time routing mechanism that moves tokens between layers, improving efficiency without omitting necessary computations.
Alongside diffusion transformers, state‑space models (SSMs) have emerged as promising alternatives to DiTs \cite{hu2024zigma, fei2024scalablediffusionmodelsstate, fei2024diffusion, teng2024dimdiffusionmambaefficient} to mitigate quadratic complexity. Explorations of token pruning \cite{zhan2024exploringtokenpruningvision}, token masking \cite{tang2024mambamimpretrainingmambastate} and mixture‑of‑experts \cite{pioro2024moemambaefficientselectivestate} for SSMs have not targeted diffusion backbones, whereas we consider token routing in SSMs aswell to boost overall training efficiency and convergence.

\vspace{-3mm}
\paragraph{Token-based Routing, Pruning and Masking.}
Several recent methods~\cite{sun2024ecditscalingdiffusiontransformers, FeiDiTMoE2024, park2024switchdiffusiontransformersynergizing} utilize Mixture-of-Experts (MoE)~\cite{jacobs1991moe, shazeer2017outrageouslylargeneuralnetworks} to improve the efficiency of diffusion transformers. Typically, MoE is implemented as a router module that divides parts of the network, such as single FFNs, into parallel parts, called ``experts". The router then determines which expert processes each token, thereby reducing computational overhead compared to applying every parameter to every token. Another strategy, Mixture-of-Depths (MoD)~\cite{raposo2024mixture}, involves a router module that decides the computation paths for each token within the network, allowing tokens to skip certain computational blocks. MoD has been shown to decrease runtime costs while maintaining model performance in transformer-based language models. In contrast to these methods, our approach does not require a routing module. Instead, we introduce a training strategy that enhances efficiency without modifying the base architecture.

To further address the scalability issues associated with attention, numerous studies~\cite{rao2021dynamicvit, meng2022adavit, fayyaz2022ats, wang2024attentiondriventrainingfreeefficiencyenhancement} have focused on pruning tokens based on their content, such as similarity or redundancy. This token pruning helps reduce the computational burden of the attention mechanism by eliminating less important tokens. While our approach selects tokens, they are skipping the computations in certain layers, it differs from prior works in that our routing scheme makes these tokens available again in deeper parts of the network and does not rely on the content of the tokens. Instead of pruning based on token content, our method employs a routing mechanism that directs tokens through the network irrespective of their content, maintaining computational efficiency.

In addition to pruning-based methods, recent advancements like MaskDiT~\cite{zheng2023fast_maskdit, Gao_2023_ICCV} have proposed token masking schemes applicable to DiTs. MaskDiT improves efficiency by significantly lowering the cost per iteration and accelerating training while matching the performance of standard DiTs. Based on this, SD-DiT~\cite{zhu2024sddit} enhances generative quality by incorporating a discriminative loss alongside the masking strategy. Similar to these approaches, our scheme directs tokens to later layers at random. However, unlike MaskDiT, which replaces masked tokens with learnable embeddings, our method routes tokens back to later layers in the network.

\begin{figure}[t]
  \centering
  \adjustbox{max width=\linewidth}{
  \includegraphics[width=\linewidth]{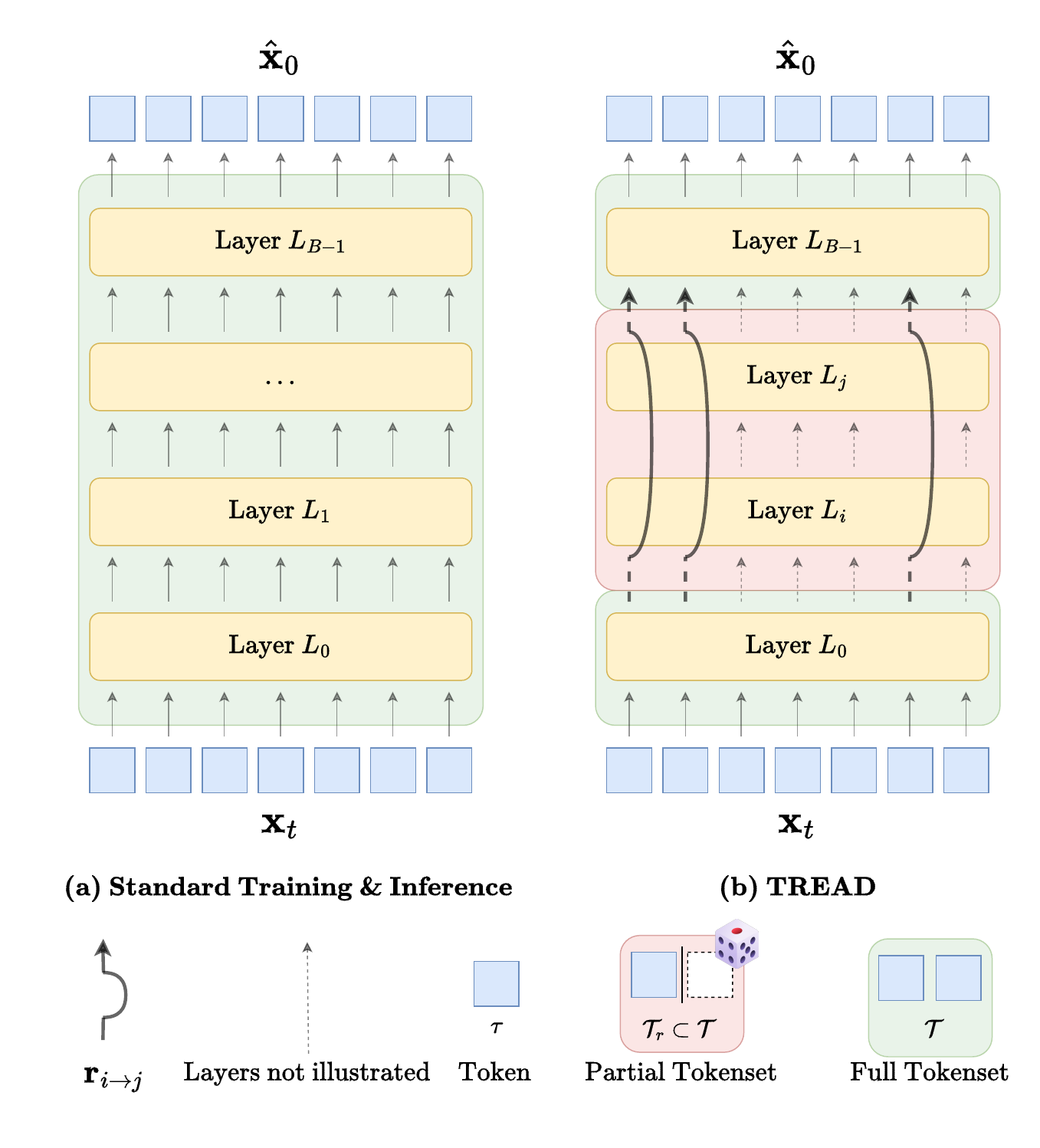}
  }
  \caption{\textbf{\raisebox{-0.1em}{\emojione[][0.3cm]} \methodname: Our method for efficient diffusion training. }
  In a) the standard training and inference strategy is shown where all tokens are processed by all layers of the network. \methodname{} enhances training efficiency by routing tokens around certain layers by reducing computational load and it preserves information which is shown in b). Since \methodname{} is used only during training, the standard setting shown in a) is used for inference.}
  \label{fig:tread_single_route}
\end{figure}

\section{\raisebox{-0.2em}{\emojione[][0.5cm]} \methodname}
\label{sec:method}

We adopt the continuous-time diffusion framework proposed by Karras et al.~\cite{karras2022elucidating} (EDM), where the forward diffusion process transforms real data samples  into progressively noisier versions through the addition of Gaussian noise:
\begin{equation}\label{eq:forward_sde_simplified}
\mathbf{x}_t = \mathbf{x}_0 + \mathbf{n}, \quad \mathbf{n} \sim \mathcal{N}(0, t^2 \mathbf{I}),
\end{equation}
where $t$ is the diffusion time.

The denoising network  estimates the original data by minimizing the denoising diffusion objective:
\begin{equation}\label{eq:diff_objective}
\mathbb{E}_{\mathbf{x}_0 \sim p_{\text{data}}} \mathbb{E}_{\mathbf{n} \sim \mathcal{N}(0, t^2 \mathbf{I})} \left\| D_\theta(\mathbf{x}_0 + \mathbf{n}, t) - \mathbf{x}_0 \right\|^2.
\end{equation}
To efficiently model global context, we represent inputs ($x_t$) through \textit{tokens}, which are continuous vector embeddings of image patches or latent segments. Tokenization allows architectures such as DiT~\cite{dit_peebles2022scalable} or SSMs \cite{hu2024zigma, peng2023rwkv} to capture long-range dependencies via global mixers. 

\paragraph{Balancing Cost and Effectiveness.}
Training diffusion models is challenging due to their sample inefficiency and high computational demands. Two primary factors must be balanced during optimization: \emph{training cost} (iterations per second) and \emph{training effectiveness} (performance gain per iteration). Increasing iterations per second reduces the cost per iteration, while improved training effectiveness lowers the overall compute time by achieving higher quality outputs in fewer steps. Many methods optimize one factor at the expense of the other, creating a trade-off between cost and effectiveness. For example, masking a subset of tokens~\cite{zheng2023fast_maskdit} significantly enhances throughput in transformer networks, which are hindered by quadratic complexity $\mathcal{O}(n^2)$, and similar benefits are observed in SSMs~\cite{li2024videomamba}. Alternatively, training effectiveness is sometimes enhanced via architectural changes (e.g., additional parameters~\cite{zheng2023fast_maskdit, tang2024generative}) or pretrained teacher models for representation distillation~\cite{yu2024repa}. However, these architecture modifications generally compromise training efficiency as more parameters or additional models are required during training, whereas masking-based methods that enhance training effectiveness~\cite{devlin2018bert, he2022masked_mae, Gao_2023_ICCV} are limited by the inherent drawback of permanent data loss.

Although masking can be used to trade training efficiency for effectiveness (or vice versa) \cite{zheng2023fast_maskdit, Gao_2023_ICCV}, it violates the core principle of diffusion, which relies on small, iterative denoising steps. Masking completely removes tokens from the information flow, requiring the model to reconstruct information solely from learnable substitutes \cite{he2022masked_mae, Gao_2023_ICCV}, thereby partially disrupting the process of incremental refinement. Despite this, the fact that the use of masking yields valid diffusion models demonstrates that it is still possible to learn the data distribution with a subset of tokens.

\vspace{-1.5mm}
\paragraph{Routing: Information‑Preserving Token Transport.}
To resolve the trade-off between training efficiency and training effectiveness, we introduce \methodname{}, a diffusion training strategy that preserves information and, therefore aligns with the core diffusion principle of incremental denoising steps. We achieve this through \emph{routing}, where a subset of tokens is temporarily removed at a \emph{start layer} and reintroduced at a later \emph{end layer}, enabling token transport without permanently discarding information. This is presented in \Cref{fig:tread_single_route}.
For standard token-based diffusion models, the denoiser is represented as a composition of layers:
\begin{equation}
D_\theta(\mathbf{x}) \equiv [L_{B-1} \circ L_{B-2} \circ \cdots \circ L_0](\mathbf{x}),
\end{equation}
where $B$ denotes the total number of layers and each token $\tau_k \in \mathcal{T}$ is processed by all layers. In contrast, our routing strategy introduces an alternate pathway that allows a subset of tokens to bypass several layers. Specifically, we define a route $\route_{i\rightarrow j}$ (with start layer $L_i$ and end layer $L_j$) as a bypass of layers $L_i, \ldots, L_j$ that is applied to a subset of tokens $\mathcal{T}_{\route_{i \rightarrow j}} \subset \mathcal{T}$. This results in the routed denoiser
\begin{equation}
    D_{\theta}^{\route_{i\rightarrow j}} = L_{B-1} \circ \cdots \circ \left\{
    \begin{array}{ll}
        \!\!\!\operatorname{id}, & \!\!\tau_k \in \mathcal{T}_{\route_{i \rightarrow j}}\!\!\!\!\\[1mm]
        \!\!\! L_j \circ \cdots \circ L_i, & \!\!\text{otherwise}\!\!\!\!
    \end{array}
    \right\} \circ \cdots \circ L_0,
\end{equation}
where $\operatorname{id}(\cdot)$ denotes the identity mapping. The routed tokens $\mathcal{T}_{\route_{i \rightarrow j}}$ are randomly selected for each training sample using a selection rate (see \Cref{tab:mask_ratios}). In the special case of $\route_{0\rightarrow j}$, the routed tokens correspond to the network input.

We integrate routing into the objective \Cref{eq:diff_objective} as
\begin{equation}
    \mathcal{L} = \mathbb{E}_{\mathbf{x}_0 \sim p_{\text{data}}} \, \mathbb{E}_{\mathbf{n} \sim \mathcal{N}(0, t^2 \mathbf{I})} \Bigl\| \network^{\route_{i\rightarrow j}}\bigl(\mathbf{x}_0 + \mathbf{n}, t\bigr) - \mathbf{x}_0 \Bigr\|^2.
\end{equation}
Unlike masking-based methods—which typically apply a standard diffusion objective to unmasked tokens and a MAE loss to masked tokens to compensate for the irreversible information loss introduced by masking~\cite{he2022masked_mae}—our approach maintains the simplicity of diffusion models by employing a single reconstruction loss uniformly across all tokens.

\paragraph{Training Efficiency.}
By skipping the computation of a large number of layers for a substantial fraction of tokens during training, our routing strategy significantly reduces computational cost. Beyond the general linear reduction in computational cost incurred in blocks such as feedforward networks, token routing also enables quadratic gains for self-attention layers. This leads to substantial reductions in computational cost for even small rates of tokens selected for routing. This efficiency gain increases with the route length. Therefore, for maximum computational efficiency, route length should be optimized.

\begin{figure}[b]
    \centering
    \includegraphics[width=1.\linewidth]{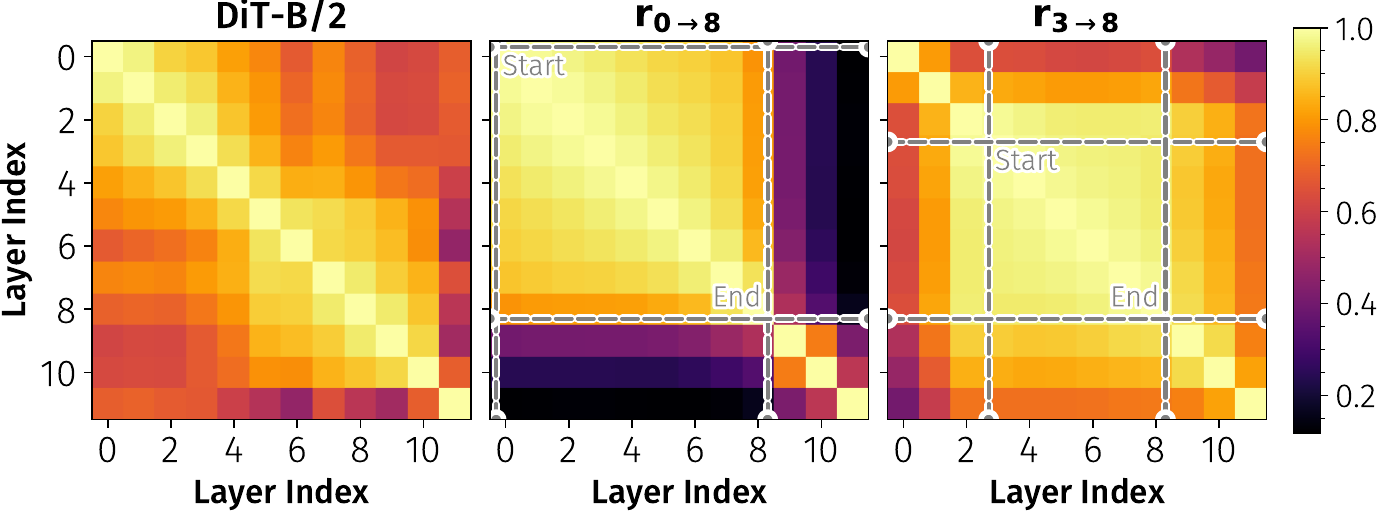}
    \caption{\textbf{Consecutive layers have highly similar output.} The effects of the routing mechanism are evident in the cosine similarities between layers. For $\route_{3\rightarrow8}$, $L_2$ exhibits high similarity with the routed layers. This is interpreted as an adaptation of $L_2$ to $\route_{3\rightarrow8}$.
    }
    \label{fig:cossim}
\end{figure}

\paragraph{Training Effectiveness.}\label{sec:effective}
Residual-based architectures, including transformers~\cite{resnet}, can interpret outputs from preceding layers~\cite{raposo2024mixture, ma2023deepcache, he2024matters, ma2024learningtocacheacceleratingdiffusiontransformer}. We demonstrate this property in \Cref{fig:cossim}, where the cosine similarity between the outputs of all layers in a trained network is shown. This characteristic can be leveraged either to cache layer outputs from previous timesteps for improved inference or to route tokens from one layer to deeper layers, as proposed in \methodname{}. While diffusion models are typically trained using the same sequence of layers for all samples for each denoising step, \methodname{} introduces additional variance. Specifically, the second pathway via route $\route_{i \rightarrow j}$ reduces the number of transformations that are applied to all tokens by the route length. As tokens are randomly selected for routing, the network learns the denoising step with a variable number of layers. This implicitly motivates the network to produce less washed-out features in layer $i$ as they are later being used by layer $j$ directly. Since the network always has access to the standard full pathway (the selection rate for $\route_{i \rightarrow j}$ is less than 1.0), it is beneficial to design the route such that the alternative pathway differs clearly from the standard one as this will increase the above mentioned effect. Good generative performance and low FID (see \Cref{fig:analysis_plots}) also correlate strongly with route length. Consequently, longer routes improve both training efficiency and training effectiveness, yielding multiplicative speed-ups (iterations per second $\times$ effectiveness per iteration).

\section{Experiment}\label{sec:experiments}
\subsection{Experimental Details}
\paragraph{Model Architecture.}
Our model follows the standard two-stage training process of Latent Diffusion Models (LDM) \cite{rombach2022high_latentdiffusion_ldm}, where an autoencoder is first trained to map between pixel-space and a compressed latent representation. We utilize the pre-trained VAE from Stable Diffusion with a standard downsampling factor of 8. As our method primarily requires token-based processing, we adopt a standard DiT-XL architecture \cite{dit_peebles2022scalable} with a patch size of 2 as our main model. Additionally, we demonstrate the generalizability of our approach to other architectures, such as Diffusion-RWKV \cite{fei2024diffusion} and Mamba \cite{gu2023mamba}. For ablations we use a DiT-B/2 if not otherwise specified. All modifications are explicitly stated for each experiment.
To indicate the application of our method, we append +\methodname{} to the respective models.

\vspace{-2mm}
\paragraph{Training Setup.}
We follow the standard of existing works on diffusion models and train models on ImageNet-256 with a batch size of $256$. Furthermore, we follow the standard setting for AdamW from the original DiT~\cite{dit_peebles2022scalable} with a learning rate of 1e-4 and no weight decay and the standard $\beta$-parameters of $0.9$ and $0.999$ respectively. We use the standard EMA approach with an update rate of $0.9999$. All results are computed using this EMA model. We do not employ any kind of data augmentation. All experiments are conducted on $4 \times 80$GB-A100 GPUs with a local batch size of 64, resulting in a global batch size of 256, which reflects the standard for ImageNet-256.
In addition, we use precomputed latent representations of ImageNet-256, leading to a $32 \times 32$ representation. 
Except for experiments that specifically state otherwise, we apply a selection rate of $50\%$ to our routing mechanism. 
For Transformer-based structures, we use random selection, while for sequential models like RWKV and Mamba, we find that row selection works better.

\paragraph{Evaluation.}
Following standard evaluation protocols \cite{dit_peebles2022scalable, yu2024repa, zheng2023fast_maskdit}, we primarily evaluate the quality of generated samples using the Fr\'echet Inception Distance (FID) \cite{heusel2017gans_fid}, reporting values on $50{,}000$ samples unless stated otherwise. To ensure fair comparisons with prior works and baselines, we adopt the standard ADM evaluation suite \cite{dhariwal2021diffusion}. Additionally, we report the sliced Fr\'echet Inception Distance (sFID) \cite{ding2022continuous}, Inception Score \cite{salimans2016improved}, as well as Precision and Recall \cite{sajjadi2018assessing}.

\subsection{Main Result}

\begin{table}[h!]
    \centering
    \normalsize
    \adjustbox{max width=\linewidth}{
    \begin{tabular}{
        l@{\hskip 5pt}
        c@{\hskip 5pt}
        c@{\hskip 5pt}
        c@{\hskip 5pt}
        c@{\hskip 5.8pt}
        c@{\hskip 5pt}
        c@{\hskip 5pt}
    }
    \toprule
    \textbf{Method}
    & \textbf{Epochs}
    & \textbf{FID}$\downarrow$
    & \textbf{sFID}$\downarrow$
    & \textbf{IS}$\uparrow$
    & \textbf{Prec.}$\uparrow$
    & \textbf{Rec.}$\uparrow$ \\
    \midrule
    \multicolumn{7}{l}{\textit{Unguided}} \\
    \rowcolor{gray!8}
    SD-DiT-XL/2 \cite{zhu2024sddit}$^\dagger$
        & 480
        & 7.21
        & 5.17
        & 144.68
        & 0.72
        & 0.61 \\
    MDT-XL/2 \cite{Gao_2023_ICCV}$^\dagger$
        & 1300
        & 6.23
        & 5.23
        & 143.02
        & 0.71
        & 0.65 \\
    \rowcolor{gray!8}
    SiT-XL/2 \cite{ma2024sitexploringflowdiffusionbased}
        & 1400
        & 8.61
        & 6.32
        & 131.65
        & 0.68
        & 0.67 \\
    MaskDiT-XL/2 \cite{zheng2023fast_maskdit}$^\dagger$
        & 1600
        & 5.69
        & 10.34
        & 177.99
        & 0.74
        & 0.60 \\
    \rowcolor{gray!8}
    DiT-XL/2$_\text{+REPA}$\cite{yu2024repa}$^\dagger$
        & 170
        & 9.60
        & -
        & -
        & -
        & -    \\
    DiT-XL/2 \cite{dit_peebles2022scalable}
        & 1400
        & 9.62
        & 6.85
        & 121.50
        & 0.67
        & 0.67 \\
    \rowcolor{gray!8}
    \hspace{1em}\textbf{+TREAD}
        & 160
        & 4.96
        & 6.83
        & 192.49
        & 0.79
        & 0.54 \\
    \hspace{1em}\textbf{+TREAD}
        & 680
        & \textbf{3.93}
        & 7.32
        & 211.35
        & 0.76
        & 0.60\\
    \midrule
    \multicolumn{7}{l}{\textit{Guided}} \\
    \rowcolor{gray!8}
    SD-DiT-XL/2 \cite{zhu2024sddit}$^\dagger$
        & 480
        & 3.23
        & -
        & -
        & -
        & -    \\
    MDT-XL/2 \cite{Gao_2023_ICCV}$^\dagger$
        & 1300
        & 1.79
        & 4.57
        & 283.01
        & 0.81
        & 0.61 \\
    \rowcolor{gray!8}
    MaskDiT-XL/2 \cite{zheng2023fast_maskdit}$^\dagger$
        & 1600
        & 2.28
        & 5.67
        & 276.56
        & 0.80
        & 0.61 \\
    SiT-XL/2 \cite{ma2024sitexploringflowdiffusionbased}
        & 1400
        & 2.06
        & 4.50
        & 270.30
        & 0.82
        & 0.59 \\
    \rowcolor{gray!8}
    DiT-XL/2 \cite{dit_peebles2022scalable}
        & 1400
        & 2.27
        & 4.60
        & 278.24
        & 0.83
        & 0.57 \\
    \hspace{1em}\textbf{+TREAD}(F)
        & 220
        & 2.76
        & 4.61
        & 252.78
        & 0.83
        & 0.57 \\
    \rowcolor{gray!8}
    \hspace{1em}\textbf{+TREAD}(F)
        & 740
        & 2.09
        & 4.46
        & 269.07
        & 0.81
        & 0.62 \\
    \hspace{1em}\textbf{+TREAD}(F)$^\ddagger$
        & 740
        & \textbf{1.69}
        & 4.73
        & 292.66
        & 0.81
        & 0.63 \\
    \bottomrule
    \end{tabular}
    }
    {\fontsize{6}{7}\selectfont ($\dagger$) alters the architecture or uses additional parameters.}
    \ \ {\fontsize{6}{7}\selectfont ($\ddagger$) uses interval guidance \cite{kynkäänniemi2024applyingguidancelimitedinterval}.}
    \caption{\textbf{Performance comparison on ImageNet-256.} Applying \methodname{} to the standard DiT results in \textit{faster convergence} while being \textit{substantially cheaper} to train. \methodname{} outperforms various baselines including those with additional parameters and architectural changes. Selected samples from \methodnameoutsidetab{DiT}{XL/2}{}{}(F) can be found in \Cref{fig:samples}.}
    \label{tab:extensive}
\end{table}

\begin{figure}[t]
    \centering
    \normalsize
    \begin{subfigure}{0.49\linewidth}
        \includegraphics[width=\linewidth]{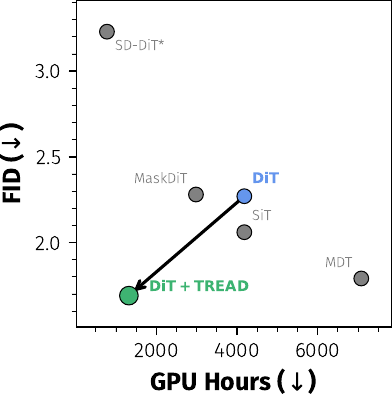} 
        \caption*{\hspace{7mm} Best FID (guided).}
        \label{fig:c}
    \end{subfigure}
    \hfill
    \begin{subfigure}{0.49\linewidth}
        \includegraphics[width=\linewidth]{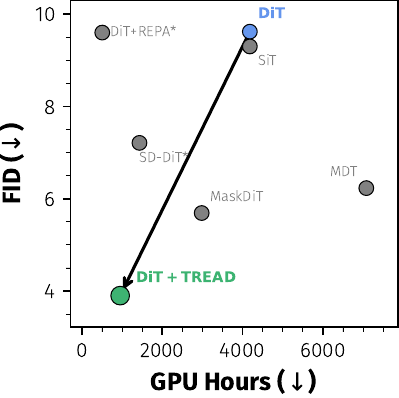} 
        \caption*{\hspace{7mm} Best FID (unguided).}
        \label{fig:b}
    \end{subfigure}
    \caption{\textbf{\methodname{} shows good performance at low compute cost.} We demonstrate a strictly better performance-cost trade-off than all other presented methods, including DiT+REPA~\cite{yu2024repa}. For methods with ($^*$), we assume an identical iteration speed to the DiT. This is advantageous for our competitors as those utilize additional parameters or entire pre-trained vision encoders to aid their diffusion model, effectively decreasing their iteration speed.}
    \label{fig:three_fid_vs_gpu_hours}
    \vspace{-5mm}
\end{figure}

\paragraph{General Comparison.} We conduct a comprehensive evaluation against state-of-the-art approaches that focus on enhancing \textit{training effectiveness}, such as MDT \cite{Gao_2023_ICCV} and SD-DiT \cite{zhu2024sddit}, as well as methods that prioritize \textit{efficiency} by accelerating iterative training, such as MaskDiT \cite{zheng2023fast_maskdit}. Our results show that \methodname{} outperforms the standard DiT \cite{dit_peebles2022scalable} in both aspects, achieving faster convergence while maintaining high efficiency and also outperforms state-of-the-art baselines in their respective domain. We refer readers to \Cref{fig:fid_steps_scatter} for a visual representation of our model's superior performance.
Specifically, w.g.t. \textit{training effectiveness}, our approach significantly outperforms both MDT and SD-DiT while we achieve higher \textit{training efficiency} through a increased iteration rate ($3.03$it/s), thus reducing computational costs compared to MaskDiT ($2.98$it/s), the baseline DiT ($1.86$it/s), and MDT ($1.02$it/s) under identical conditions.
Our efficiency is also shown in \Cref{tab:extensive}, where \methodname{} demonstrates accelerated convergence and superior performance. In the unguided setting, our approach not only accelerates training but also improves performance while reducing computational costs (see \Cref{fig:three_fid_vs_gpu_hours}).
In the guided setting, we first train the model with \methodname{} and then fine-tune it without routing to improve its response to classifier-free guidance (CFG). This follows the well-documented phenomenon that restricting information during training can degrade performance under CFG \cite{zheng2023fast_maskdit, zhu2024sddit}. While our method outperforms prior work and the standard DiT without fine-tuning in the guided setting (see Appendix \Cref{fig:fid_cfg_dit_b_comparison}), we still observe an improved CFG response after a subsequent fine-tuning stage without routing. To this end, we adopt the fine-tuning strategy from \citet{zheng2023fast_maskdit}, using a batch size of 1{,}024 and a reduced learning rate for 75K iterations.

\vspace{-4mm}
\paragraph{Comparison with Base Architectures.}
We demonstrate the effectiveness of our approach in improving efficiency and performance while showcasing its adaptability to diverse architectures. To illustrate this, we use RWKV~\cite{peng2023rwkv}, which shares many aspects of SSMs and employs a RNN-like structure, which makes it fundamentally different from the transformer-based DiT.
As shown in \Cref{tab:dit}, our method consistently outperforms the baseline models across both architectures. Notably, the performance gains are also pronounced in larger models, indicating the scalability of our approach and its suitability for larger architectures. We also expand \methodname{} beyond EDM \cite{karras2022elucidating} to flowmatching \cite{lipman2022flow, ma2024sitexploringflowdiffusionbased, albergo2022building} and find that our method transfers well to other frameworks. All models, except those denoted as ``SiT'' in \Cref{tab:dit} and \Cref{tab:custom_baseline} are trained using EDM to remain comparable to our baselines and competitors.
We also extend our experiments beyond ImageNet-256 to ImageNet-512 to investigate the impact of larger token counts (see \Cref{tab:imgnet512mscoco_results}).
\begin{table}[t]
\centering
\normalsize
\adjustbox{max width=.9\linewidth}{
\begin{tabular}{l c c c l}
  \toprule
  \textbf{Model}                          & \textbf{\#Param} & \textbf{Iter.} & \textbf{Batchsize} & \multicolumn{1}{c}{\textbf{FID} $\downarrow$}      \\
  \midrule

  DiT-S/2 \cite{dit_peebles2022scalable}  & 33M             & 400K           & 256              & \fidbaseline{68.40}                 \\
  \rowcolor{gray!8}
  \hspace{1em}\textbf{+ \methodname}      & 33M             & 400K           & 256              & \fidimproved{\textbf{47.04}}{21.36} \\
  SiT-S/2 \cite{ma2024sit}                & 33M             & 400K           & 256              & \fidbaseline{57.60}                 \\
  \rowcolor{gray!8}
  \hspace{1em}\textbf{+ \methodname}      & 33M             & 400K           & 256              & \fidimproved{\textbf{38.08}}{19.52} \\
  \midrule

  DiT-B/2 \cite{dit_peebles2022scalable}  & 130M            & 400K           & 256              & \fidbaseline{43.47}                 \\
  \rowcolor{gray!8}
  \hspace{1em}\textbf{+ \methodname}      & 130M            & 400K           & 256              & \fidimproved{\textbf{21.34}}{22.13} \\
  SiT-B/2 \cite{ma2024sit}                & 130M            & 400K           & 256              & \fidbaseline{33.00}                 \\
  \rowcolor{gray!8}
  \hspace{1em}\textbf{+ \methodname}      & 130M            & 400K           & 256              & \fidimproved{\textbf{15.98}}{16.02} \\
  \midrule

  DiT-L/2 \cite{dit_peebles2022scalable}  & 458M            & 400K           & 256              & \fidbaseline{23.33}                 \\
  \rowcolor{gray!8}
  \hspace{1em}\textbf{+ \methodname}      & 458M            & 400K           & 256              & \fidimproved{\textbf{\phantom{0}9.10}}{14.23}  \\
  SiT-L/2 \cite{ma2024sit}                & 458M            & 400K           & 256              & \fidbaseline{18.80}                 \\
  \rowcolor{gray!8}
  \hspace{1em}\textbf{+ \methodname}      & 458M            & 400K           & 256              & \fidimproved{\textbf{\phantom{0}5.91}}{12.89}  \\
  \midrule

  DiT-XL/2 \cite{dit_peebles2022scalable} & 675M            & 400K           & 256              & \fidbaseline{19.47}                 \\
  \rowcolor{gray!8}
  \hspace{1em}\textbf{+ \methodname}      & 675M            & 400K           & 256              & \fidimproved{\textbf{\phantom{0}7.38}}{12.09}  \\
  SiT-XL/2 \cite{ma2024sit}               & 675M            & 400K           & 256              & \fidbaseline{17.20}                 \\
  \rowcolor{gray!8}
  \hspace{1em}\textbf{+ \methodname}      & 675M            & 400K           & 256              & \fidimproved{\textbf{\phantom{0}4.89}}{12.31}  \\
  \midrule
  LightningDiT \cite{yao2025reconstruction} & 675M & 80K        & 1024              & \fidbaseline{\phantom{0}5.14}                 \\
  \rowcolor{gray!8}
  \hspace{1em}\textbf{+ \methodname}      & 675M            & 80K           & 1024              & \fidimproved{\textbf{\phantom{0}3.71}}{1.43}  \\
  \midrule

  RWKV \cite{fei2024diffusion}            & 210M            & 400K           & 256              & \fidbaseline{59.84}                 \\
  \rowcolor{gray!8}
  \hspace{1em}\textbf{+ \methodname}      & 210M            & 400K           & 256              & \fidimproved{\textbf{53.79}}{6.05}  \\
  Mamba \cite{maml}                       & 111M            & 400K           & 256              & \fidbaseline{69.39}                 \\
  \rowcolor{gray!8}
  \hspace{1em}\textbf{+ \methodname}      & 111M            & 400K           & 256              & \fidimproved{\textbf{61.17}}{8.22}  \\

  \bottomrule
\end{tabular}
}
\caption{\textbf{Performance increase across all model configurations} trained up to benchmark setting on ImageNet-256 (400K / 80K iterations and batch size 256 / 1024).}
\label{tab:dit}
\end{table}

\begin{table}[b]
\centering
\normalsize
\adjustbox{max width=0.6\linewidth}{
    \begin{tabular}{lcl}
        \toprule
        \textbf{Model} & \textbf{Iter.} & \multicolumn{1}{c}{\textbf{FID} $\downarrow$} \\
        \midrule
        DiT-B/2 &  & 43.47 \\
        \rowcolor{gray!8}
        \hspace{1em}+ REPA &  & 34.63 \\
        \hspace{1em}+ \textbf{\methodname} &  & 21.34 \\ 
        \rowcolor{gray!8}
        \hspace{1em}+ \textbf{\methodname} + REPA & \multirow{-4}{*}{400K} & \fidimproved{\textbf{20.06}}{1.28} \\
        \midrule
        \multicolumn{3}{l}{\textit{Further Training + Unguided}} \\
        \rowcolor{gray!8}
        \hspace{1em}+ REPA &  & 28.86 \\
        \hspace{1em}+ \textbf{\methodname} &  & 15.87 \\
        \rowcolor{gray!8}
        \hspace{1em}+ \textbf{\methodname} + REPA & \multirow{-3}{*}{800K} & \fidimproved{\textbf{13.76}}{2.37} \\
        \multicolumn{3}{l}{\textit{Further Training + Guided}} \\
        \rowcolor{gray!8}
        \hspace{1em}+ REPA &  & \phantom{0}9.04 \\
        \hspace{1em}+ \textbf{\methodname} &  & \phantom{0}6.39 \\
        \rowcolor{gray!8}
        \hspace{1em}+ \textbf{\methodname} + REPA & \multirow{-3}{*}{800K} & \phantom{0}\fidimproved{\textbf{5.62}}{0.77} \\
        \bottomrule
    \end{tabular}
}
\caption{\textbf{Combining the \methodname{} training strategy with REPA} to demonstrate that our method's modular design enables combination with orthogonal approaches to further improve performance.}
\label{tab:tread_p_repa}
\end{table}

\vspace{-5mm}
\paragraph{Compounding Gains with Representation Distillation.}
As \methodname{} does not modify the base architecture and, unlike masking-based methods, does not incur information loss, it is naturally orthogonal to other changes in diffusion training and can thus be combined with them. We evaluate this by combining \methodname{} with REPA~\cite{yu2024repa}, a state-of-the-art method that incorporates representation distillation from a pretrained vision backbone~\cite{oquab2023dinov2} as a secondary objective to improve diffusion models.
As shown in \Cref{tab:tread_p_repa}, the combination of both methods results in better generation performance (as indicated by a lower FID) than each achieves individually. The reduced costs and improved effectiveness through \methodname{} enable compounding gains when combined with other diffusion training methods.

\begin{figure}[t]
    \centering
    \normalsize
    \begin{subfigure}{0.24\linewidth}
        \includegraphics[width=\linewidth]{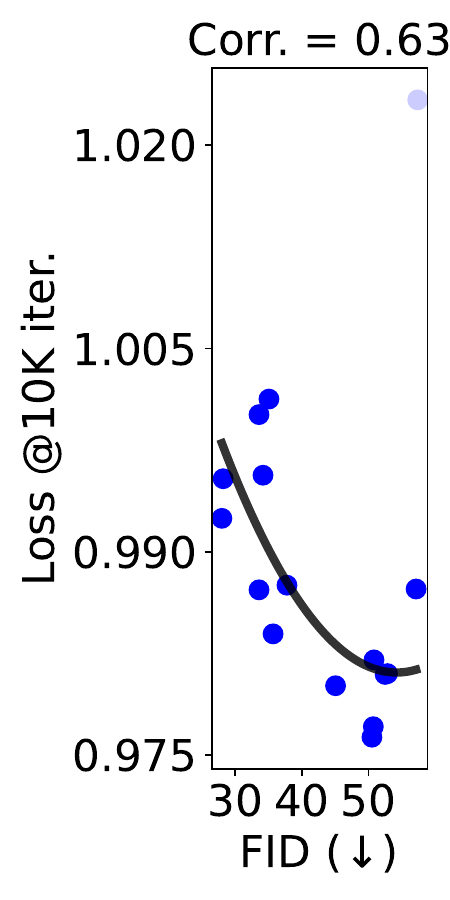}
    \end{subfigure}
    \hfill
    \begin{subfigure}{0.49\linewidth}
        \includegraphics[width=\linewidth]{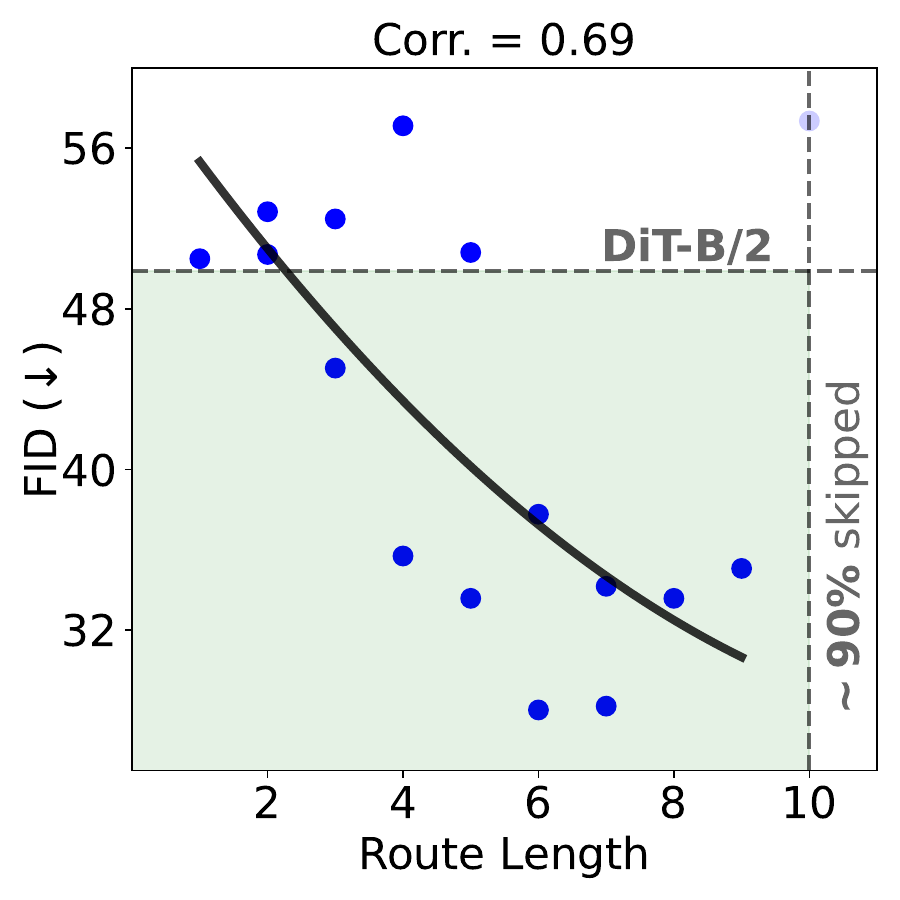}
    \end{subfigure}
    \hfill
    \begin{subfigure}{0.24\linewidth}
        \includegraphics[width=\linewidth]{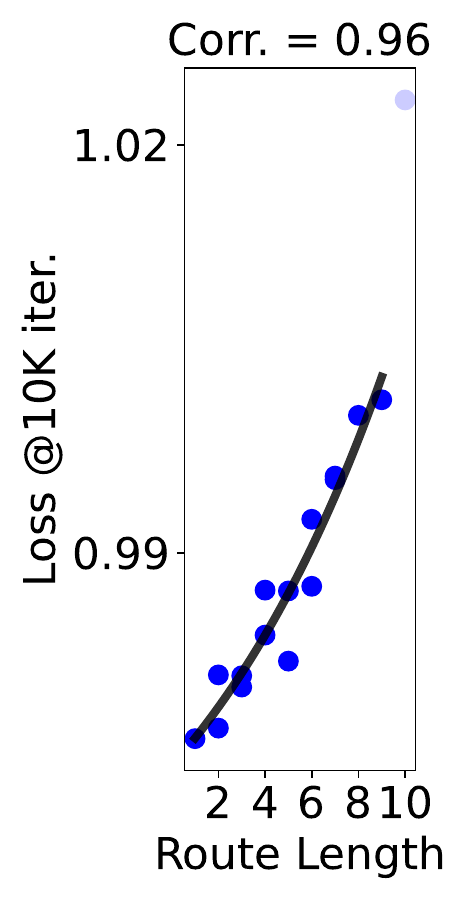}
    \end{subfigure}
    \caption{\textbf{Early training loss strongly correlates with FID improvements.} Our analysis indicates that route length shows a strong correlation with both the FID and early training loss. Consequently, the observed correlation between training loss and FID can be attributed to route length acting as a hidden variable.}
    \label{fig:analysis_plots}
    \vspace{-2.5mm}
\end{figure}

\vspace{-3.5mm}
\paragraph{Scaling to Larger Resolutions and Text-to-Image.}
We also investigate the scalability and general applicability of \methodname{} by validating it in higher-resolution and more general settings. Specifically, we evaluate on ImageNet-512 and T2I MS-COCO (2014)~\cite{coco}. We train DiT-B/2 models with and without \methodname{} for 400K iterations each using a batchsize of 256 and evaluate using FID@10k for both. We find that our method leads to substantial gains in generative performance, reducing FID by more than a third (as indicated in \Cref{tab:imgnet512mscoco_results}), while also leading to substantially faster training due to a significant increase in iteration speed.

\begin{table}[b]
  \centering
  \normalsize
  \vspace{-2mm}
  \adjustbox{max width=0.9\linewidth}{
    \begin{tabular}{lcccl}
      \toprule
      \textbf{Model} & \textbf{\#Param} & \textbf{Iter.} & \textbf{Batchsize} & \multicolumn{1}{c}{\textbf{FID@10K} $\downarrow$} \\
      \midrule
      \multicolumn{5}{l}{\textit{ImageNet-512}} \\
      DiT                                  & 130M & 400K & 256 & 62.88 \\
      \rowcolor{gray!8}
      \hspace{1em}+ \textbf{\methodname}  & 130M & 400K & 256 & \fidimproved{\textbf{40.91}}{21.97} \\
      \midrule
      \multicolumn{5}{l}{\textit{MS-COCO}} \\
      DiT                                  & 130M & 400K & 256 & 35.68 \\
      \rowcolor{gray!8}
      \hspace{1em}+ \textbf{\methodname}  & 130M & 400K & 256 & \fidimproved{\textbf{20.55}}{15.13} \\
      \bottomrule
    \end{tabular}
  }
  \vspace{-2mm}
  \caption{\textbf{Application of \methodname{} to high-resolution class conditional generation and text-to-image generation.} We demonstrate \methodname{}'s effectiveness is transferable to other generative diffusion tasks like high-resolution ImageNet-512 and T2I MS-COCO. For both experiments we use a DiT-B/2 and \methodnameoutsidetab{DiT}{B/2}{}{}.}
  \label{tab:imgnet512mscoco_results}
\end{table}

\subsection{Ablation Study}

\begin{figure}[t]
    \centering
    \includegraphics[width=\linewidth]{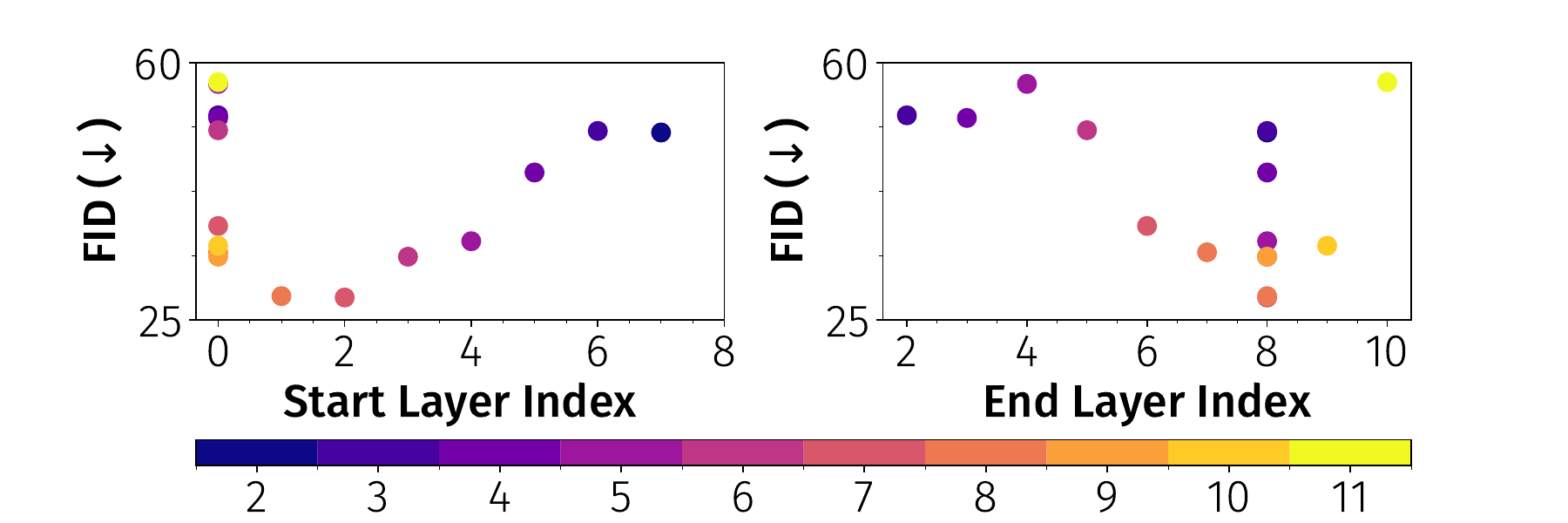}
    \caption{\textbf{Routes should start early and end in time.} This results in longer routes and allows the model to spend reasonable remaining capacity to reintegrate the routed tokens. This is clearly demonstrated as later start indices ($i$) and earlier ending indices ($j$) correlate with worse generative performance (measured with FID). }
    \label{fig:slope_chart}
    \vspace{-4mm}
\end{figure}

We conduct multiple ablation studies to examine the effect that each component has on the performance of \methodname. All experiments are conducted using a DiT-B/2 model and evaluated using $50{,}000$ samples if not otherwise specified.

\begin{figure*}[t]
    \centering
    \includegraphics[width=\linewidth]{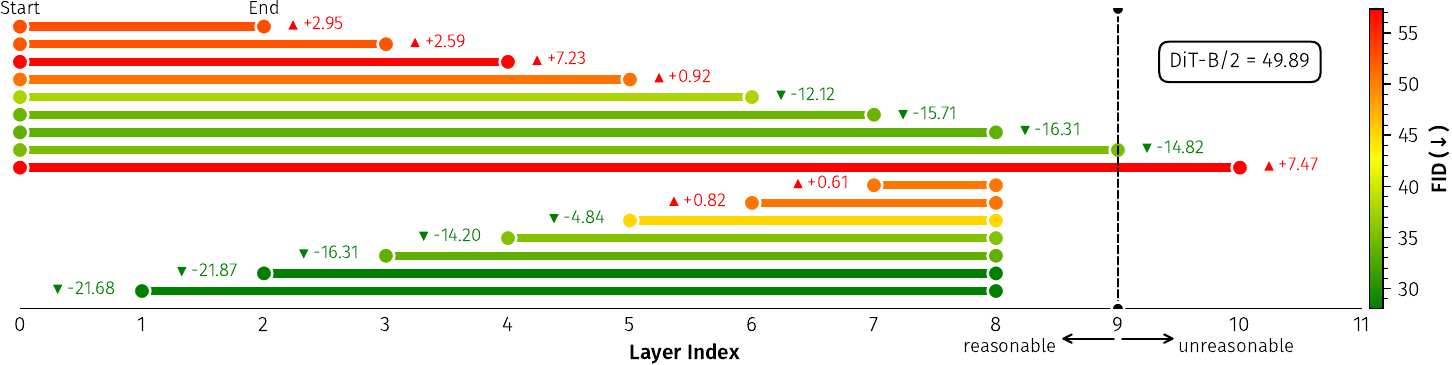}
    \caption{\textbf{Longer routes lead to better performance.}  
    We evaluate DiT-B/2 models trained for 400K iterations using FID@5K.  
    The results show that FID improves as the route length increases. The lowest FID values are associated with longer routes, early start layers, and relatively late end layers. However, we find that there exists a breakpoint when end layers are chosen too late, as the network requires enough capacity to incorporate routed tokens.}
    \label{fig:route_ablation}
    \vspace{-1mm}
\end{figure*}

\vspace{-3mm}
\paragraph{Route Location.}
In \Cref{fig:route_ablation}, we present results from a set of DiT-B/2 models trained with different routes $\route_{i\rightarrow j}$. We conduct two ablations: first, by fixing $i=0$ and varying the end layer $j$, and second, by fixing $j=8$ and varying the starting layer $i$. 
From these experiments, we observe two distinct modes: one where routing improves performance and one where it does not. Since these modes are clearly separable, we can derive principles for route placement to enhance training effectiveness. Specifically, we find that \textbf{i)} longer routes generally lead to better performance, \textbf{ii)} starting layers up to $i=2$ improve results, and \textbf{iii)} ending layers up to $j=8$ are beneficial. Additionally, the configuration \(\route_{i=0, j=10}\) demonstrates that choosing an excessively late end location \(j\) can be detrimental to performance, as the network requires sufficient capacity to integrate routed tokens into the overall information flow. Whenever routing does not lead to lower FID, we can attribute the observed effect to violating at least one of these principles.
In the case of \methodname{}, we can optimize both training effectiveness and training efficiency simultaneously. According to our route placement guidelines, training effectiveness can be increased with longer routes. At the same time, longer routes accelerate training speed by increasing the number of layers processing fewer tokens compared to DiT \cite{dit_peebles2022scalable}.
In \Cref{fig:analysis_plots}, we present our findings from this analysis. Our results indicate that route length is strongly correlated with both FID and training loss, even at early stages. Consequently, early training loss also exhibits a strong correlation with FID. Additionally to the usual diffusion loss we see an offset caused by the route that indicates its impact on the model. Furthermore, we provide a custom baseline in \Cref{tab:custom_baseline} where we compare \methodname{} against token-level dropout in each layer to underline the importance of route configuration.
We also demonstrate the importance of the start layer $i$ and end layer $j$ in \Cref{fig:slope_chart}, where we show the generative performance for models with the respective route placement settings. Notably, there is a trend for both where the best results were achieved using $i=2$ and $j=8$, with deviations following a clear downwards direction in terms of FID (except for $j=10$).

\vspace{-1mm}
\paragraph{Selection Rate.}
In this experiment, we ablate the impact of the selection ratio on performance, as shown in \Cref{tab:mask_ratios}, where the selection ratio is defined in the range $\in [0, 1]$. We specifically exclude $1.0$ as it would entirely skip layers of the network, leaving substantial parts untrained. We ablate in increments of $0.25$. We observe a significant performance improvement when using a random selection rate of $0.25$ ($\text{FID} = 26.50$) and $0.5$ ($\text{FID} = 21.34$), both outperforming the vanilla DiT-B/2 ($\text{FID} = 43.47$) while also reducing computational cost per iteration. Interestingly, the medium selection rate of $0.5$ achieves a lower FID than $0.25$. However, a higher selection rate of $0.75$ ($\text{FID} = 49.29$) leads to degraded performance, performing worse than the vanilla model. This aligns with our explanation in \Cref{sec:effective}, as the selection rate affects the influence of the alternative token pathway during training. In addition, a high selection rate causes fewer tokens to be processed in each layer affected by the route, leading to better training efficiency.

\begin{table}[h]
  \centering
  \begin{minipage}[t]{0.2\textwidth}
    \centering
    \footnotesize
    \scalebox{.9}{%
      \begin{tabular}{ll}
        \toprule
        \textbf{Selection Rate} & \multicolumn{1}{c}{\textbf{FID} $\downarrow$} \\
        \midrule
        0 (DiT-B/2) & 43.47 \\
        0.25 & \fidimproved{26.50}{16.97} \\
        \textbf{0.50} & \fidimproved{\textbf{21.34}}{22.13} \\
        0.75 & \fidworsened{49.29}{5.82} \\
        \bottomrule
      \end{tabular}
    }
    \subcaption{Selection Rate Ablation.}
    \label{tab:mask_ratios}
  \end{minipage}%
  \hfill
  \begin{minipage}[t]{0.27\textwidth}
    \centering
    \footnotesize
    \scalebox{.9}{%
      \begin{tabular}{ll}
        \toprule
        \textbf{Model} & \multicolumn{1}{c}{\textbf{FID} $\downarrow$} \\
        \midrule
        SiT-B/2 \cite{ma2024sit} & 33.00\\
        \hspace{1em} TLDrop & \fidimproved{32.00}{1.00} \\
        \hspace{1em} TLDropR & \fidimproved{24.28}{8.72} \\
        \hspace{1em} \textbf{+\methodname{}} (absolute) & \fidimproved{\textbf{15.98}}{17.02}\\
        \bottomrule
      \end{tabular}
    }
    \subcaption{Token‑level dropout vs.\ TREAD.}
    \label{tab:custom_baseline}
  \end{minipage}
  \caption{\textbf{Ablation and Baseline Comparison.} \subref{tab:mask_ratios} shows the effect of selection rate on FID on \methodnameoutsidetab{DiT}{B/2}{}{}. \subref{tab:custom_baseline} compares token-level dropout over all layers (TLDrop), over layers $2 \rightarrow 8$ (TLDropR) and our long‑route \methodname. In all cases, the dropout resembles a route of length 1.}
  \label{tab:combined}
\end{table}

\begin{table}[h!]
    \centering
    \footnotesize
    \adjustbox{max width=\linewidth}{
        \begin{tabular}{l l}  
            \toprule
            \textbf{Model} & \multicolumn{1}{c}{\textbf{FID} $\downarrow$} \\ 
            \midrule
            DiT-XL/2 &  19.47 \\
            \hspace{1em} \textbf{+\methodname{}} (relative) & \fidimproved{18.81}{0.66}\\
            \hspace{1em} \textbf{+\methodname{}} (absolute) & \fidimproved{\textbf{\phantom{0}7.38}}{12.09}\\
            \bottomrule
        \end{tabular}
    }
    \caption{\textbf{Location choice transfers from DiT-B/2 to large models like DiT-XL/2.} We extend our insights from \Cref{fig:route_ablation} to larger models by using relative settings ($i=\frac{1}{6}$, $j=\frac{2}{3}$) and absolute settings ($i=2$, $j=B-4$) based on the optimal configuration of DiT-B/2. We find the absolute setting to achieve the best performance.}
    \label{tab:extrapolation}
    \vspace{-4mm}
\end{table}

\vspace{-5mm}
\paragraph{Extrapolating Route Location to Large Models.}
For the B/2 model, we find the optimal route to be \(\route_{i=2 \rightarrow j=8}\) as shown in \Cref{fig:route_ablation}.
To extend these findings to larger models with additional layers, two strategies are considered:
\emph{Relative Extrapolation}: In this approach, the indices are scaled proportionally to the total number of layers, resulting in
\(i = \frac{2}{12} = \frac{1}{6}\) and \(j = \frac{8}{12} = \frac{2}{3}\).
\emph{Absolute Configuration}: Here, the starting index is fixed at \(i = 2\) and the ending index is chosen as
\(j = B - 4\), where $B$ represents the total number of layers.
These approaches aim to determine whether the optimal routing configuration identified in the B/2 model can be effectively generalized to larger models. As shown in \Cref{tab:extrapolation}, the \methodnameoutsidetab{DiT}{XL/2}{}{} model trained using the absolute extrapolation approach achieves superior performance. This substantial performance gap suggests that the optimal route location scales in an absolute, or near-absolute, manner with larger model sizes.

\section{Conclusion}
Training diffusion models remains computationally expensive, even as recent methods leverage accelerated inference, fine-tuning, and personalization. We propose \methodname{}, a diffusion training strategy that integrates into token-based sequential models without architectural modifications. By routing tokens from early to later layers, \methodname{} improves generation quality while reducing computational cost at the same time. This leads to a total speed-up to standard training of \textbf{37$\times$}. We systematically explore its design space, derive empirically validated guidelines, and demonstrate that it accelerates convergence on class-conditional ImageNet-256 by an order of magnitude compared to standard training. Our extensive experiments show that \methodname{} can be applied to current state-of-the-art methods, like representation distillation, to further improve training effectiveness, enabling future work in this field for a greater number of researchers.
\section*{Acknowledgement}


This project has been supported by the Federal Ministry for Economic Affairs and Energy (BMWE) within the project “NXT GEN AI METHODS - Generative Methoden für Perzeption, Prädiktion und Planung”, the project “GeniusRobot” (01IS24083) funded by the Federal Ministry of Research, Technology and Space (BMFTR), and the bidt project KLIMA-MEMES, and partially funded by the Horizon Europe project ELLIOT (101214398). The authors gratefully acknowledge the Gauss Center for Supercomputing for providing compute through the NIC on JUWELS/JUPITER at JSC and the HPC resources supplied by the NHR @ FAU Erlangen.
{
    \small
    \bibliographystyle{ieeenat_fullname}
    \bibliography{main}
}
\appendix
\onecolumn

\setcounter{page}{1}

\part*{%
  \begin{center}
    {\Large TREAD: Token Routing for Efficient Architecture‑agnostic Diffusion Training}\\[0.75em]
    {\large --Supplementary Materials--}
  \end{center}%
}
\addcontentsline{toc}{part}{Supplementary Materials}


\renewcommand{\thetable}{S\arabic{table}}
\renewcommand{\thefigure}{S\arabic{figure}}

\section{Implementation Details}
\subsection{Experimental Configuration} In contrast to DiT \cite{dit_peebles2022scalable} and MDT \cite{Gao_2023_ICCV}, which leverage the ADM framework \cite{dhariwal2021diffusion}, our experimental approach is grounded in the formulation of EDM \cite{karras2022elucidating}. Specifically, we implement EDM's preconditioning through a $\sigma$-dependent skip connection, utilizing the standard parameter settings.

This approach eliminates the necessity to train ADM's noise covariance parameterization, as required by DiT. For the inference phase, we adopt the default temporal schedule defined by: 
\begin{equation} 
t_{i < N} = \left(t_\text{max}^{\frac{1}{\rho}} + \frac{i}{N-1} \left(t_\text{min}^{\frac{1}{\rho}} - t_\text{max}^{\frac{1}{\rho}}\right)\right)^\rho, 
\end{equation} 
where the parameters are set to $N=40$, $\rho=7$, $t_\text{max}=80$, and $t_\text{min}=0.002$. Furthermore, we employ Heun’s method as the ODE solver for the sampling process. This choice has been shown to achieve FID scores comparable to those obtained with 250 DDPM steps while significantly reducing the number of required steps \cite{zheng2023fast_maskdit, karras2022elucidating}.

The noise distribution adheres to the EDM configuration, defined by: 
\begin{equation}
\ln(p_\sigma) \sim \mathcal{N}(P_\text{mean}, P_\text{std}), 
\end{equation} 
with $P_\text{mean} = -1.2$ and $P_\text{std}=1.2$. For detailed information, refer to the EDM paper \cite{karras2022elucidating}.

\subsection{Network Details}

\paragraph{Parameter Comparison}
As previously discussed, our method does not require any modifications to the architecture itself, whereas other methods incorporate a predefined decoder head on top of the standard DiT structure. This introduces computational overhead that is particularly noticeable in smaller models. Since our method does not need these additional parameters, we reduce the computational cost associated with the decoder head. This is demonstrated in \Cref{tab:parameters}.
\begin{table}[ht!]
\centering
\begin{tabular}{lc}
\toprule
\textbf{Model} & \textbf{\# of Parameters (Millions)} \\ 
\midrule
DiT & 675 \\
\rowcolor{gray!8}$\text{DiT}_\textbf{+TREAD}$ & 675 \\
MaskDiT & 730 \\
\rowcolor{gray!8}SD-DiT & 740 \\
\bottomrule
\end{tabular}
\caption{Comparison of the number of network parameters. MaskDiT and SD-DiT add a substantial number of parameters, approximately 10\% of those in XL-sized DiT models. This additional parameter count is fixed across different model sizes \cite{zheng2023fast_maskdit, zhu2024sddit}, which can slow down smaller models since the relative size of the added decoder components increases.}
\label{tab:parameters}
\end{table}

\paragraph{Diffusion-RWKV Setting.}
Due to the nature of RWKV and other state-space models (SSMs) \cite{gu2023mamba, hu2024zigma, alkin2024visionlstmxlstmgenericvision}, a row selection strategy is applied instead of a random selection. Additionally, we adhere to the DiT configuration in the RWKV setting. Nevertheless, we are able to improve upon our own Diffusion-RWKV \cite{fei2024diffusion} baseline using \methodname{}. The poor performance of our RWKV baseline can be attributed to the number of layers; our model consists of only 12 layers, whereas \citet{fei2024diffusion} recommends using 25 or even 49 layers.

\paragraph{Mixed-Precision.}
\methodname{} can be used successfully with \texttt{bf16}. However, it is noteworthy that when less computational blocks are available towards the end (like 1-3) might run into instabilities during training. We were able to mitigate this by keeping $L_j$ in \texttt{fp32} during training when using a route $\route_{i \rightarrow j}$. The effect on iteration speed is minimal.

\subsection{Hyperparameters}
Throughout all of our experiments we use the same structure as DiT \cite{dit_peebles2022scalable}. We use AdamW \cite{loshchilov2017decoupled_adamw} and a constant learning rate of 1e-4, $(\beta_1, \beta_2) = (0.9, 0.999)$ and no weight decay. Furthermore, we train in \texttt{bf16}, precompute the data into lates using the Stable Diffusion VAE \cite{rombach2022high_latentdiffusion_ldm}. We use the \texttt{stabilityai/sd-vae-ft-ema} VAE checkpoint from huggingface.
\begin{table}[h!]
    \centering\small
    \adjustbox{max width=0.9\textwidth}{%
    \begin{tabular}{l c c c c}
        \toprule
        \rowcolor{gray!8} & DiT-S & DiT-B & DiT-L & DiT-XL \\
        \midrule
        \textbf{Optimization} \\
        Batch size & 256 & 256 & 256 & 256 \\ 
        Optimizer & AdamW & AdamW & AdamW & AdamW \\
        LR & 1e-4 & 1e-4 & 1e-4 & 1e-4 \\
        $(\beta_1, \beta_2)$ & (0.9, 0.999) & (0.9, 0.999) & (0.9, 0.999) & (0.9, 0.999) \\
        \midrule
        \textbf{Optimization - Finetune} \\
        Batch size & - & - & - & 1{,}024 \\ 
        Optimizer & - & - & - & AdamW \\
        LR & - & - & - & 1e-5 \\
        $(\beta_1, \beta_2)$ & - & - & - & (0.9, 0.999) \\
        \midrule
         \textbf{Architecture} \\
        
        Dim & 384 & 768  & 1,024 & 1,152 \\
        Heads & 6 & 12 & 16 & 16 \\ 
        Layers & 12 & 12 & 24 & 28 \\
        \midrule
        
        \midrule
        \textbf{\methodname{}} \\
        Route & $\route_{2 \rightarrow 8}$ & $\route_{2 \rightarrow 8}$ & $\route_{2 \rightarrow 20}$ & $\route_{2 \rightarrow 24}$ \\ 
        Selection Rate & 0.5 & 0.5 & 0.5 & 0.5 \\
        \bottomrule
    \end{tabular}
    }
    \caption{Hyperparameter setup for DiT variants.}
    \label{tab:hyperparam}
\end{table}

\subsection{Classifier-Free Guidance}
\begin{figure}[b]
    \centering
    \includegraphics[width=0.8\linewidth]{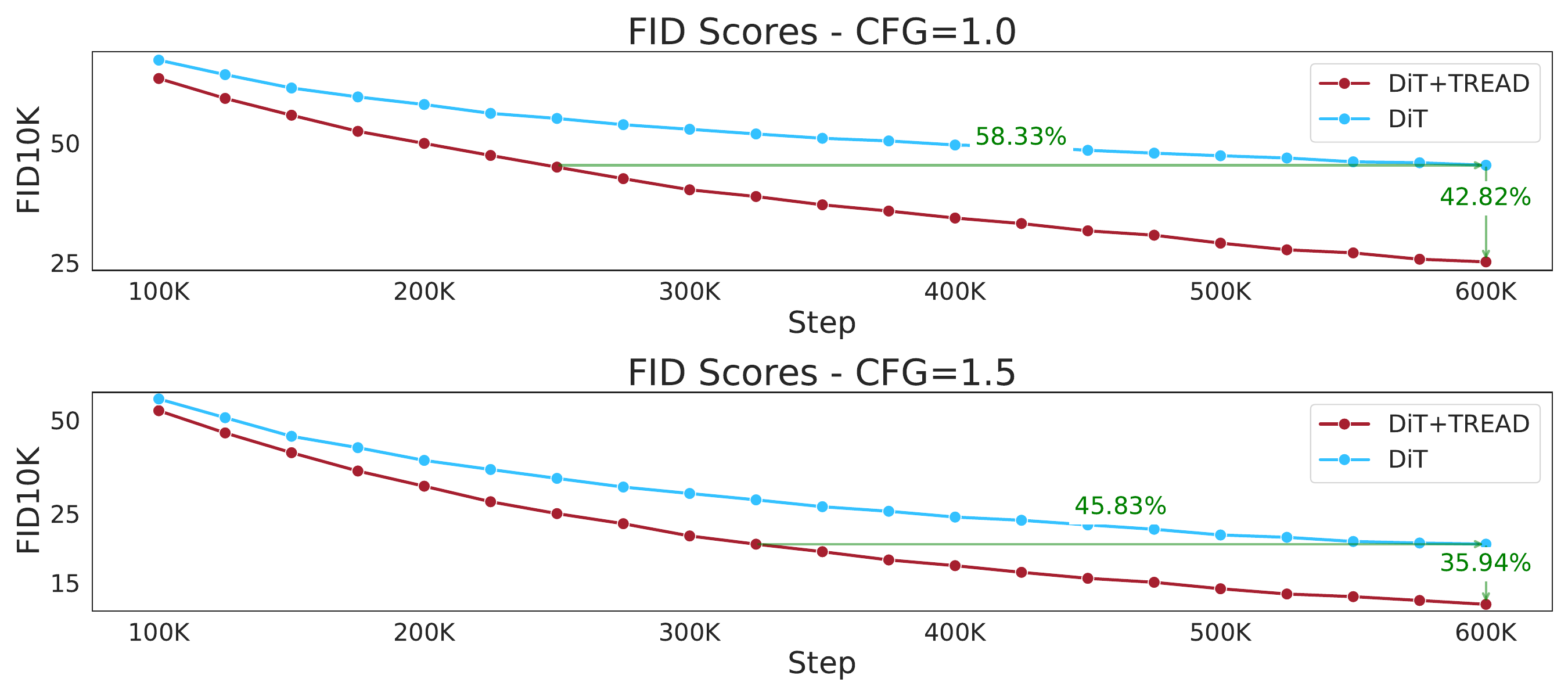}
    \caption{We compare FID@10K between a DiT-B/2 and \methodnameoutsidetab{DiT}{B/2}{}{} with and without Classifier-free Guidance (CFG). \methodname{} outperforms the standard DiT approach, even without the need to finetune without routing.}
    \label{fig:fid_cfg_dit_b_comparison}
\end{figure}

\methodname{} does demonstrate superior performance for both unguided as well as guided generation on a DiT-B/2 as shown in \Cref{fig:fid_cfg_dit_b_comparison}. 

\section{Loss Curves and Routing induced Loss Gap}
\begin{figure}[htbp]
  \centering
  \begin{subfigure}[b]{0.45\textwidth}
    \centering
    \includegraphics[width=\linewidth]{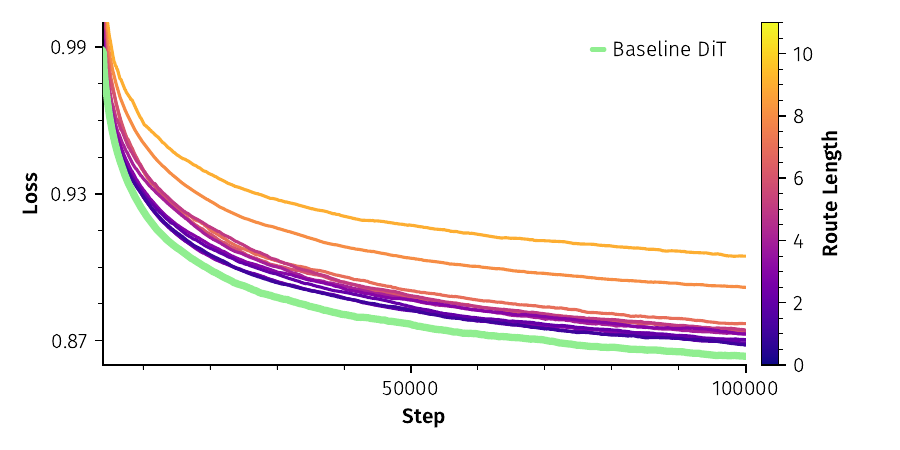}
    \label{fig:loss_curve_normal}
  \end{subfigure}
  \hfill
  \begin{subfigure}[b]{0.45\textwidth}
    \centering
    \includegraphics[width=\linewidth]{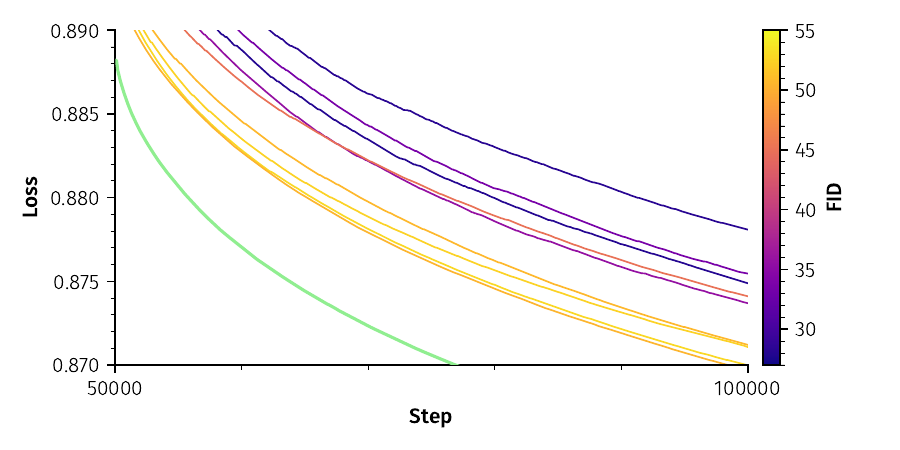}
    \label{fig:loss_curve_zoomed}
  \end{subfigure}
  \caption{\textbf{The impact of the routing mechanism on the model can be estimated with a loss difference.} We provide loss curves between 0 to 100K iterations against route length (left) and a zoomed-in version against FID (right). It can be seen that route length correlates with increased loss difference from baseline as well as with final FID.}
  \label{fig:loss_curves}
\end{figure}

We provide loss curves in \Cref{fig:loss_curves} for better understanding of the interaction between loss, route length and FID. The loss gap to the baseline DiT can be explained using the increased difficulty during training which is induced by longer routes. 

\newcommand{\horriblevspace}{\begin{tikzpicture}\node[] {};\end{tikzpicture}}
\begin{figure}
    \centering
    \includegraphics[width=\linewidth]{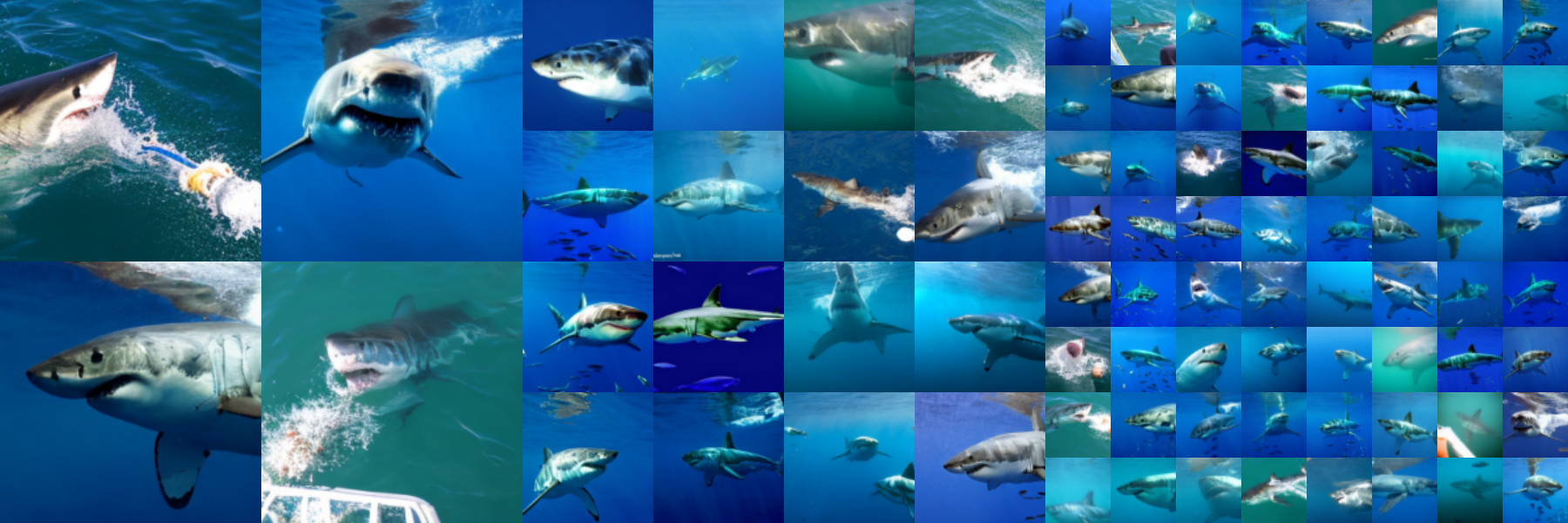}
    \horriblevspace{}
    \includegraphics[width=\linewidth]{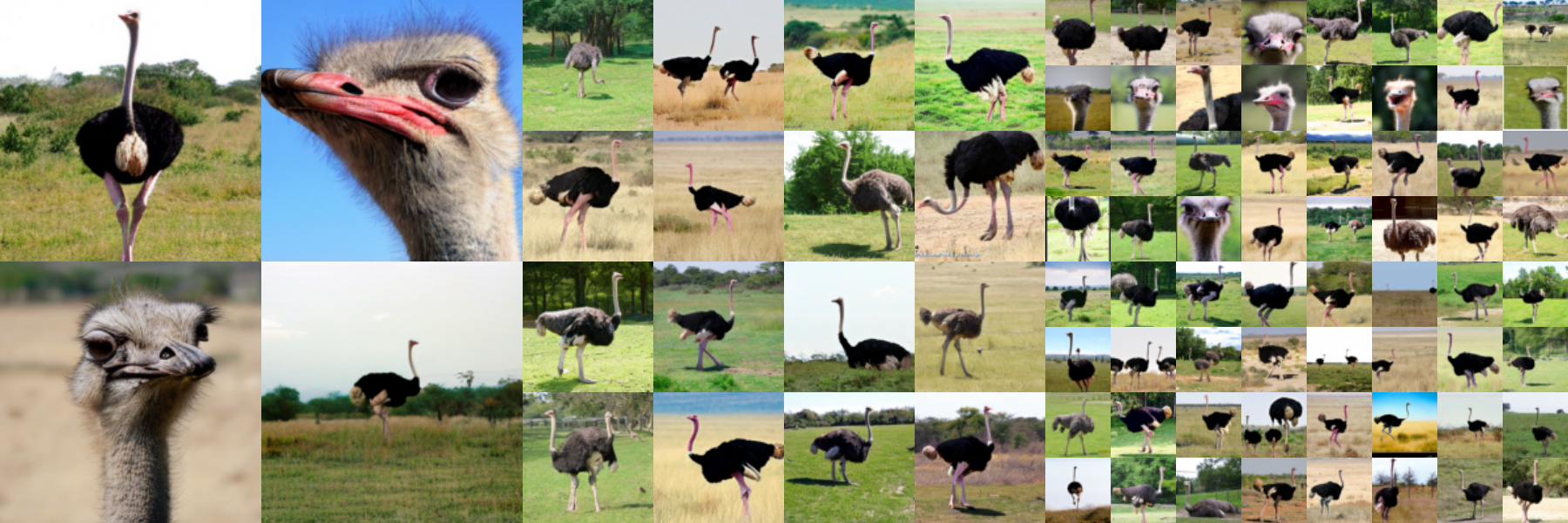}
    \horriblevspace{}
    \includegraphics[width=\linewidth]{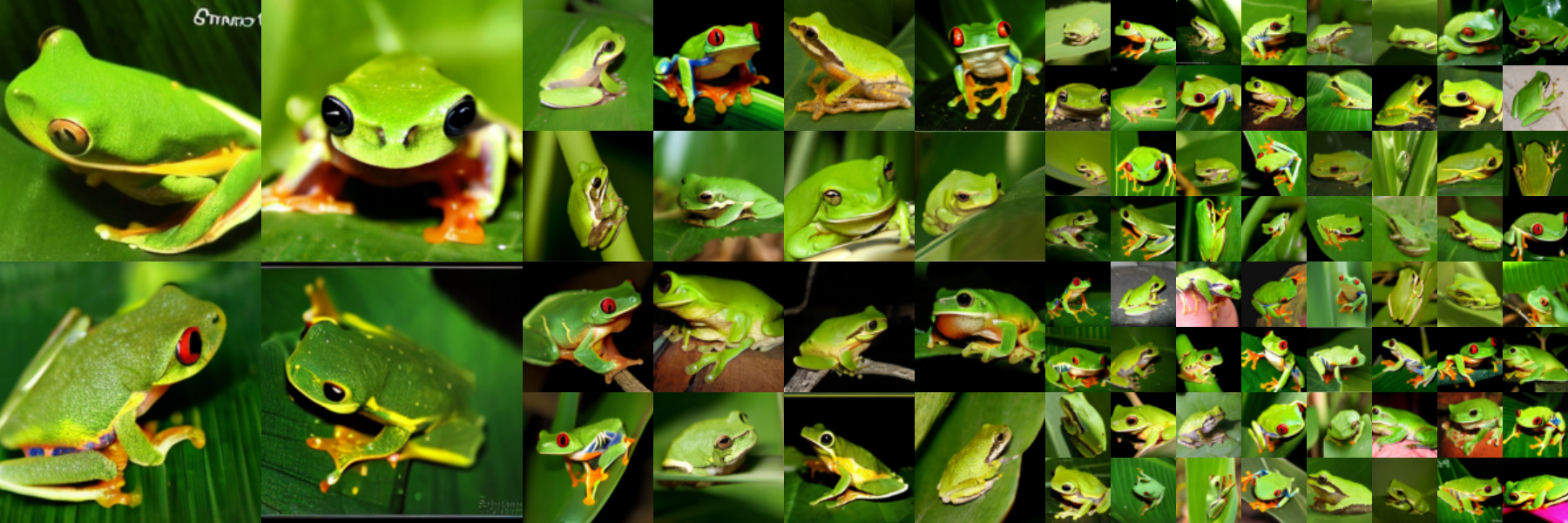}
    \caption{\textbf{Uncurated 256 $\times$ 256 samples} from \methodnameoutsidetab{DiT}{XL/2}{}{}(F) with $\omega=3.5$.}
\end{figure}

\begin{figure}
    \centering
    \includegraphics[width=\linewidth]{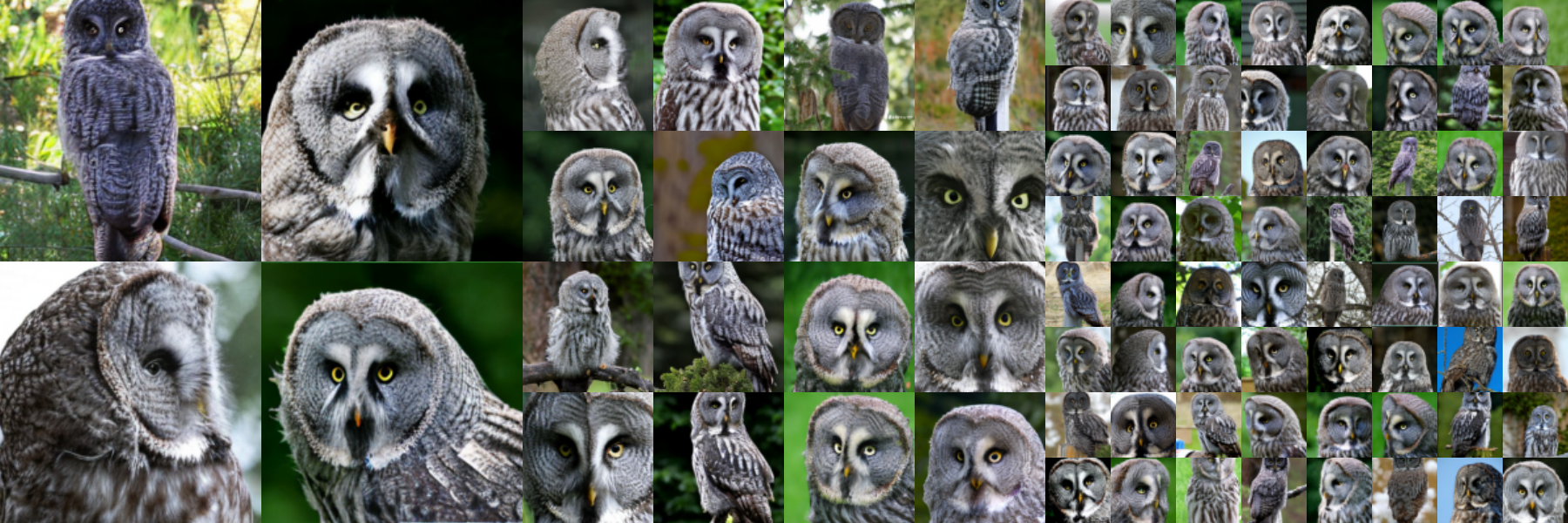}
    \horriblevspace{}
    \includegraphics[width=\linewidth]{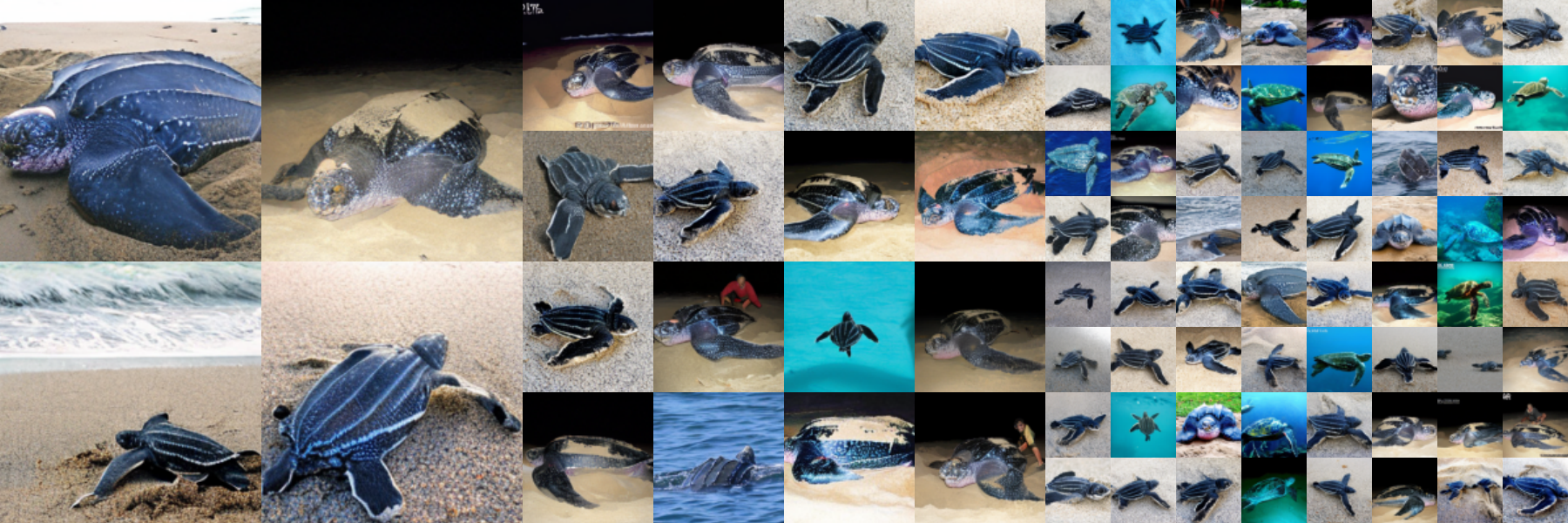}
    \horriblevspace{}
    \includegraphics[width=\linewidth]{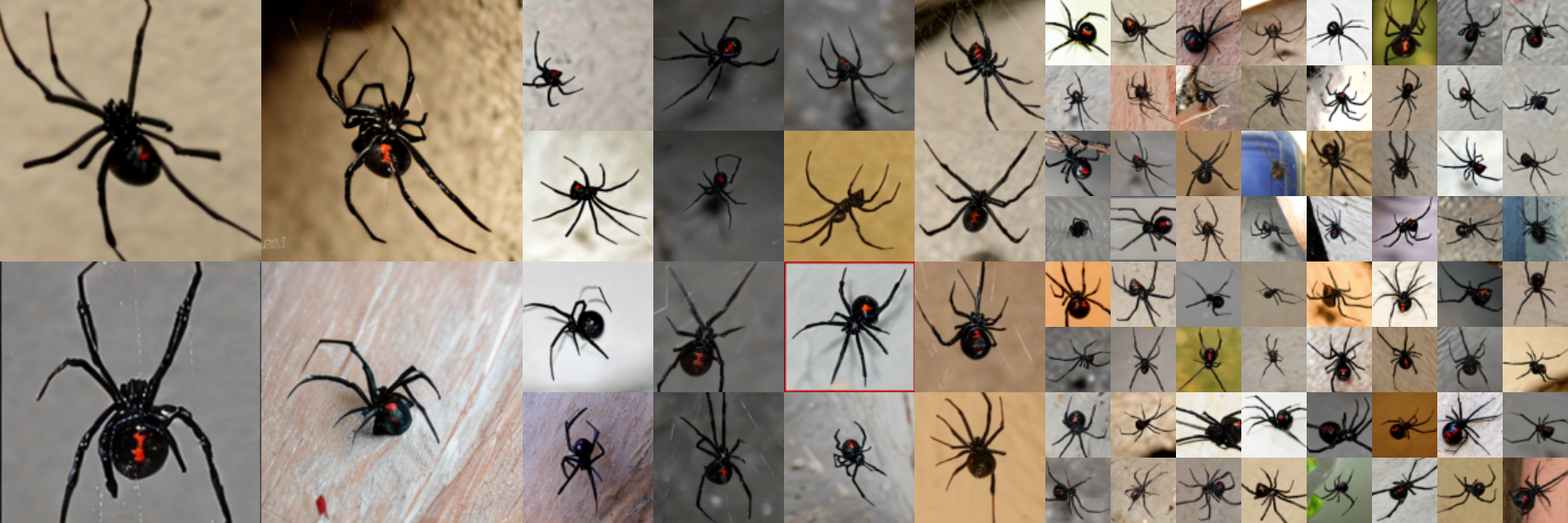}
    \caption{\textbf{Uncurated 256 $\times$ 256 samples} from \methodnameoutsidetab{DiT}{XL/2}{}{}(F) with $\omega=3.5$.}
\end{figure}

\begin{figure}
    \centering
    \includegraphics[width=\linewidth]{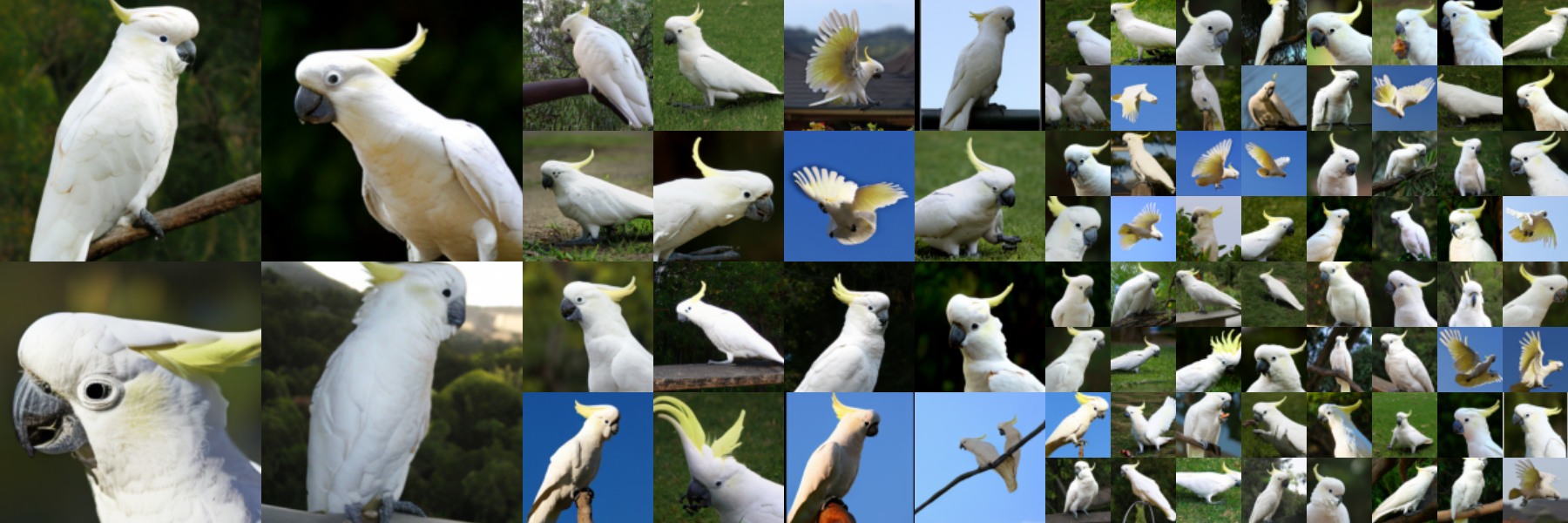}
    \horriblevspace{}
    \includegraphics[width=\linewidth]{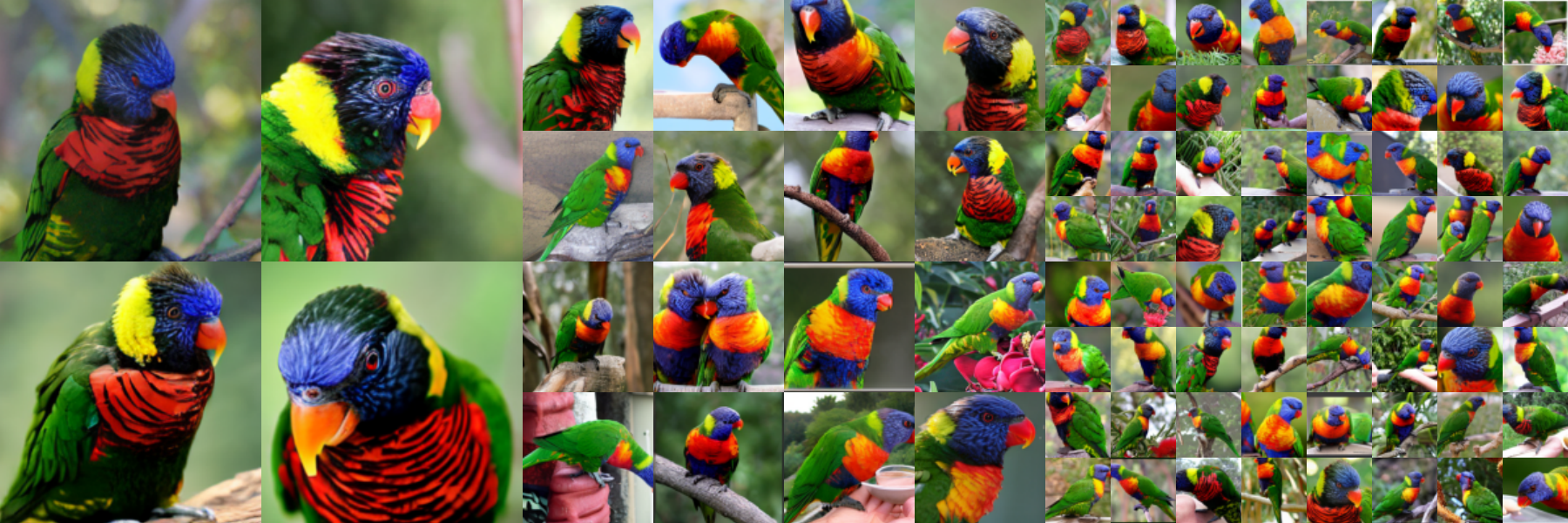}
    \horriblevspace{}
    \includegraphics[width=\linewidth]{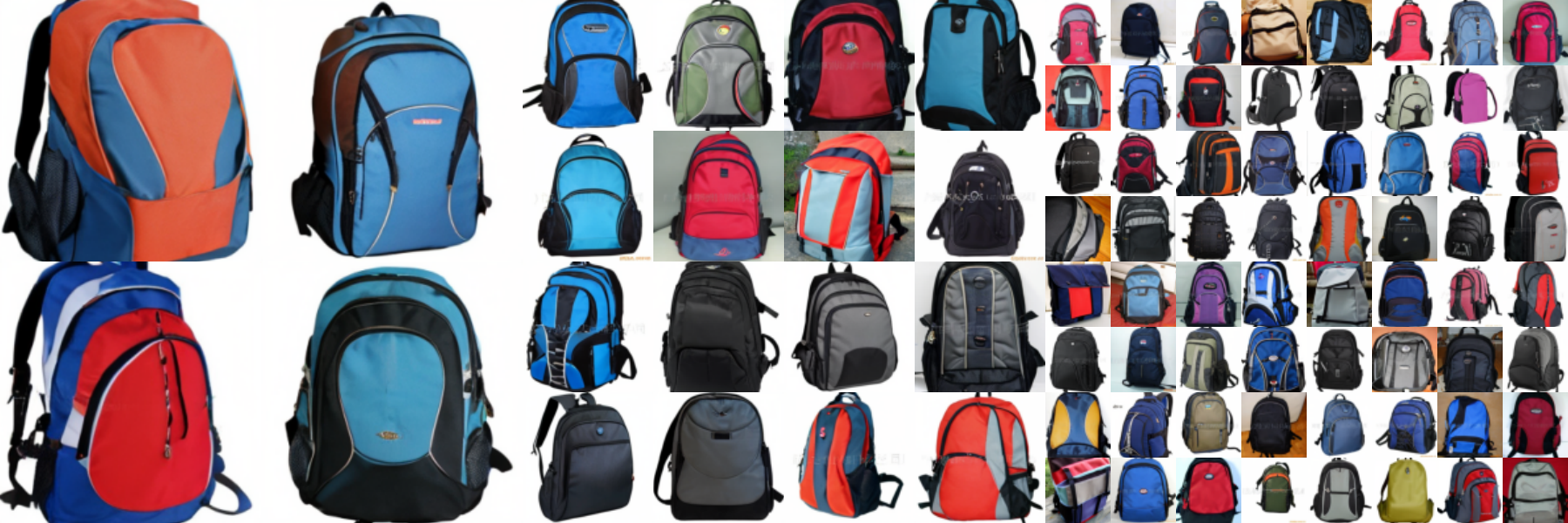}
    \caption{\textbf{Uncurated 256$\times$ 256 samples} from \methodnameoutsidetab{DiT}{XL/2}{}{}(F) with $\omega=3.5$.}
\end{figure}

\begin{figure}
    \centering
    \includegraphics[width=\linewidth]{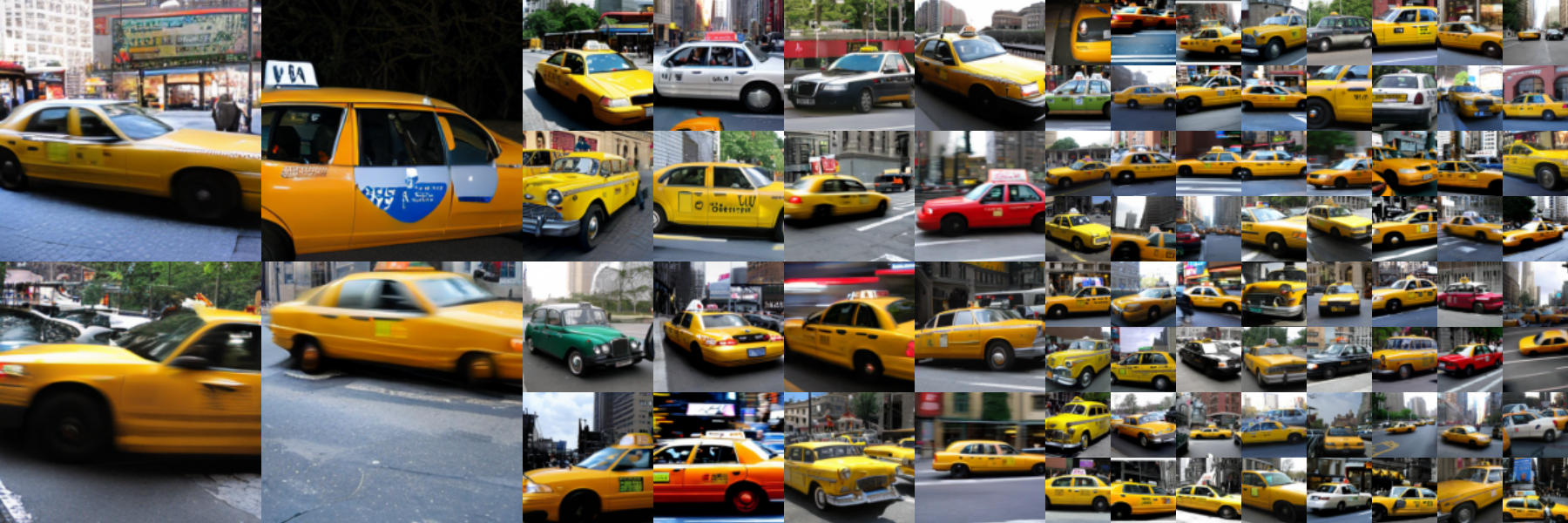}
    \horriblevspace{}
    \includegraphics[width=\linewidth]{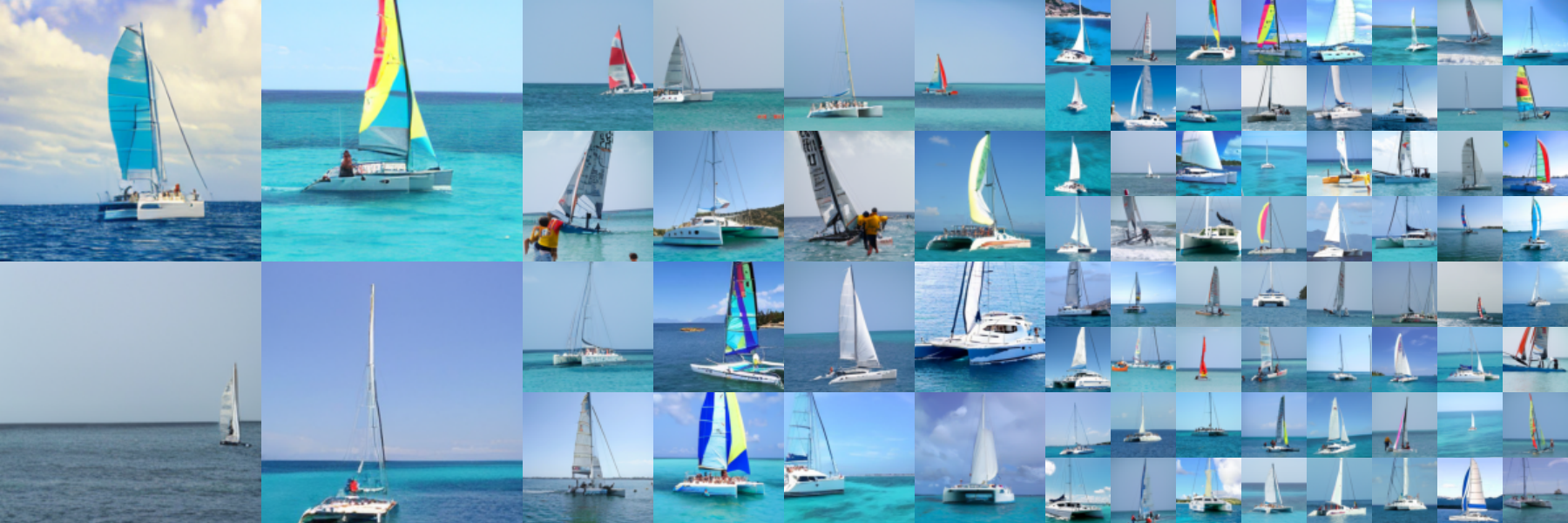}
    \horriblevspace{}
    \includegraphics[width=\linewidth]{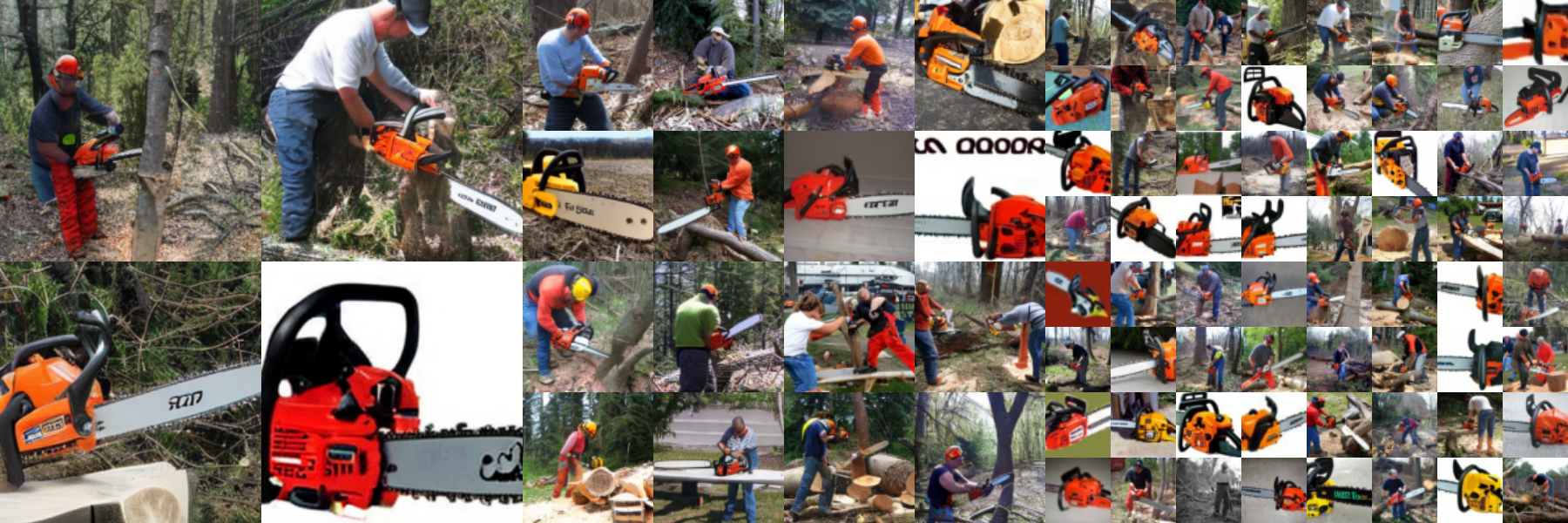}
    \caption{\textbf{Uncurated 256 $\times$ 256 samples} from \methodnameoutsidetab{DiT}{XL/2}{}{}(F) with $\omega=3.5$.}
\end{figure}

\begin{figure}
    \centering
    \includegraphics[width=\linewidth]{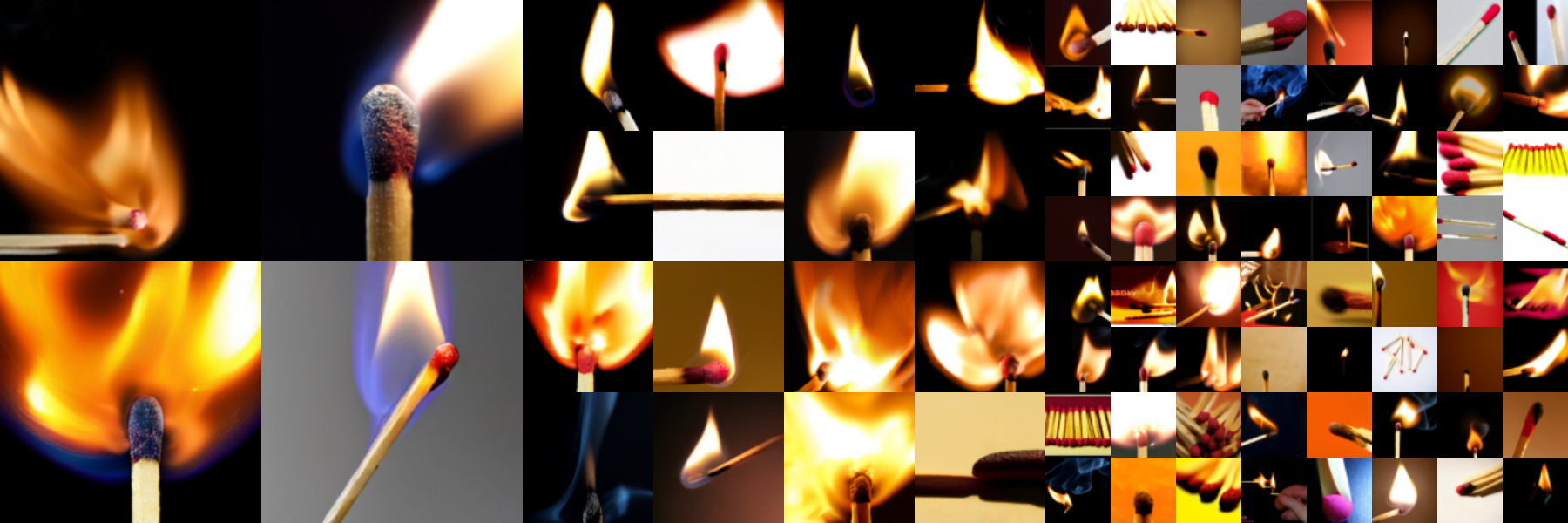}
    \horriblevspace{}
    \includegraphics[width=\linewidth]{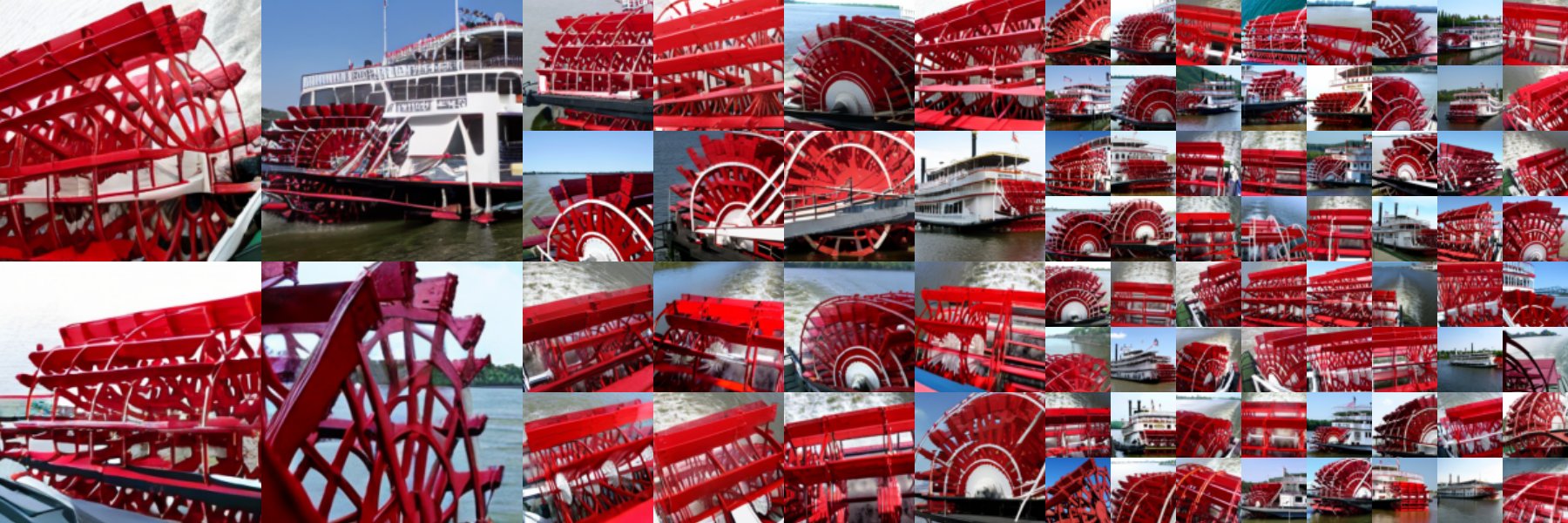}
    \horriblevspace{}
    \includegraphics[width=\linewidth]{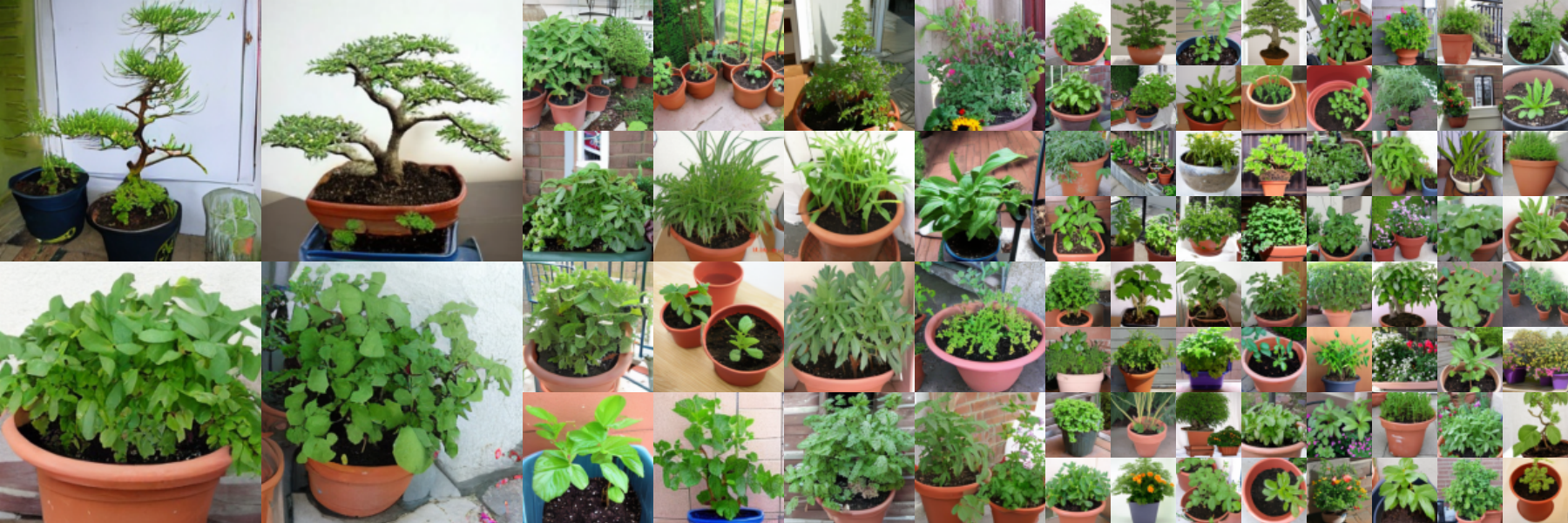}
    \caption{\textbf{Uncurated 256 $\times$ 256 samples} from \methodnameoutsidetab{DiT}{XL/2}{}{}(F) with $\omega=3.5$.}
\end{figure}

\end{document}